\documentclass{article}
\PassOptionsToPackage{numbers, compress}{natbib}

\usepackage[final]{neurips_2022}

\usepackage[utf8]{inputenc} 
\usepackage[T1]{fontenc}    
\usepackage{hyperref}       
\usepackage{url}            
\usepackage{booktabs}       
\usepackage{amsfonts}       
\usepackage{nicefrac}       
\usepackage{microtype}      
\usepackage{xcolor}         
\usepackage{natbib}
\usepackage{amssymb}
\usepackage{pifont}

\usepackage{microtype}
\usepackage{graphicx}
\usepackage{subfigure}
\usepackage{bbm}
\usepackage{booktabs} 
\usepackage{amsmath,amsfonts,amsthm,amssymb}
\usepackage{mathrsfs}

\newtheorem{definition}{Definition}
\newtheorem{proposition}{Proposition}

\newtheorem{remark}{Remark}

\usepackage{setspace}
\usepackage{graphicx}
\usepackage{subfigure}
\usepackage{hyperref}
\usepackage{xcolor}
\newcount\comments  
\newcommand{\genComment}[2]{\ifnum\comments=1{\textcolor{#1}{\textsf{\footnotesize #2}}}\fi}

\usepackage{wrapfig}
\usepackage{setspace}
\usepackage{graphicx}
\usepackage{subfigure}
\usepackage{booktabs} 
\usepackage[skip=9pt]{caption}
\DeclareCaptionFormat{myformat}{\fontsize{9}{10}\selectfont#1#2#3}
\usepackage{hyperref}
\usepackage{algorithmic}
\usepackage{algorithm}

\usepackage{microtype}
\usepackage{graphicx}
\usepackage{subfigure}
\usepackage{booktabs} 
\usepackage{setspace}
\usepackage{graphicx}
\usepackage{subfigure}
\usepackage{booktabs} 
\usepackage{microtype}
\usepackage{graphicx}
\usepackage{subfigure}
\usepackage{bm}
\usepackage{amsmath,amsfonts,amsthm,amssymb}
\usepackage{mathrsfs}

\usepackage{hyperref}

\title{Novel Policy Seeking with Constrained Optimization}

\author{%
  Hao Sun\thanks{hs789@cam.ac.uk} \\
  DAMTP, Cambridge\\
   \And
  Zhenghao Peng \\
  CS, UCLA \\
  \And 
  Bo Dai \\
  Shanghai AI Lab\\
  \And
  Dahua Lin \\
  IE, CUHK \\ 
  \And 
  Bolei Zhou \\
  CS, UCLA
}

\begin{document}

\maketitle

\begin{abstract}
    
In problem-solving, we humans tend to come up with different novel solutions to the same problem. 
However, conventional reinforcement learning algorithms ignore such a feat and only aim at producing a set of monotonous policies that maximize the cumulative reward. The resulting policies usually lack diversity and novelty. In this work, we aim at enabling the learning algorithms with the capacity of solving the task with multiple solutions through a practical novel policy generation workflow that can generate a set of diverse and well-performing policies. Specifically, we begin by introducing a new metric to evaluate the difference between policies. On top of this well-defined novelty metric, we propose to rethink the novelty-seeking problem through the lens of constrained optimization,  to address the dilemma between the task performance and the behavioral novelty in existing multi-objective optimization approaches, we then propose a practical novel policy-seeking algorithm, Interior Policy Differentiation (IPD), which is derived from the interior point method commonly known in the constrained optimization literature. Experimental comparisons on benchmark environments show IPD can achieve a substantial improvement over previous novelty-seeking methods in terms of both novelties of generated policies and their performances in the primal task. \footnote{Code is open-sourced at \href{https://github.com/holarissun/NPSCO}{https://github.com/holarissun/NPSCO}}
\end{abstract}

\section{Introduction}

In the sense of learning through interactions with the environment, the scheme of reinforcement learning (RL) is conceptually similar to the emergence of intelligence~\citep{sutton1998introduction}: an agent explores and exploits information of a given environment, learns to master some certain skills through trials and errors to gain as much reward as possible. When solving a problem, we humans could be creative to come up with multiple different solutions and gain insights from searching for diverse solutions. e.g., a self-containing example is the various approaches in RL research.

While the state-of-the-art algorithms have achieved superhuman performance in a variety of challenging tasks~\citep{vinyals2019grandmaster,akkaya2019solving,berner2019dota,badia2020agent57,elbarbari2021ltlf}, the task of encouraging individualized diversity~\footnote{Note that this is different from diversity-driven exploration. While the latter focuses on the inner-diversity of a policy, in our work we focus on the inter-diversity between policies} of learned agents, on the other hand, is relatively under-explored. Different from conventional RL agents that are only learned through interactions with the external environment, novel policy generation is a task considering the differentiation among individual policies. The differentiation among policies can be explained as the social influence~\citep{rogoff1990apprenticeship, ryan2000intrinsic, van2011social, henrich2017secret, harari2014sapiens} in social science literature. 
Although many works have been proposed applying social motivation to Multi-Agent Reinforcement Learning (MARL) settings \citep{jaques2019social, hughes2018inequity, sequeira2011emerging, peysakhovich2017consequentialist}, how to motivate a single RL agent to perform differently against existing agents is still an open question. 

\begin{figure*}[t]
\vskip 0.2in
\centering
\begin{minipage}[htbp]{0.255\linewidth}
			\centering
			\includegraphics[width=0.9\linewidth]{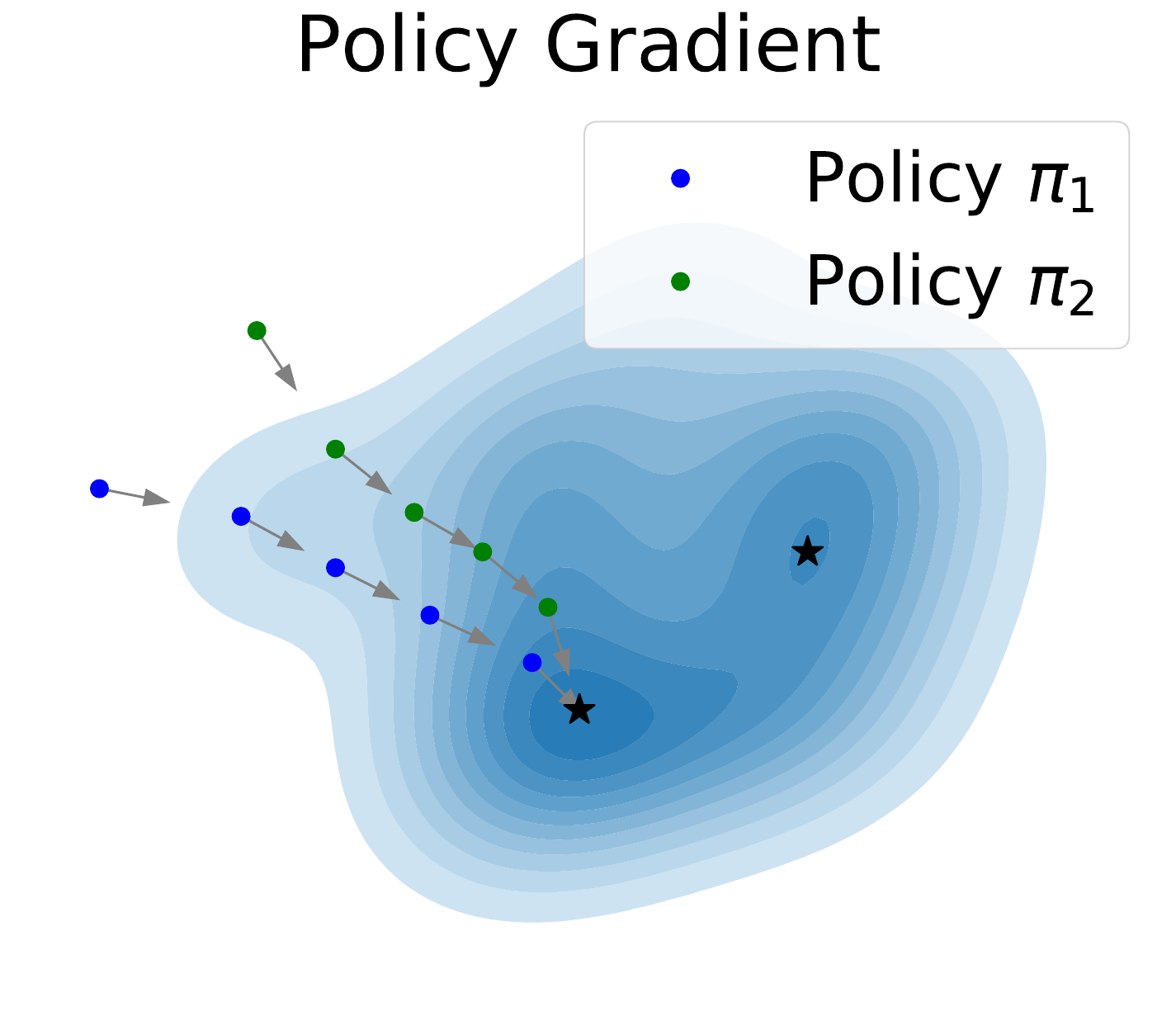}
		\end{minipage}%
		\begin{minipage}[htbp]{0.35\linewidth}
			\centering
			\includegraphics[width=0.9\linewidth]{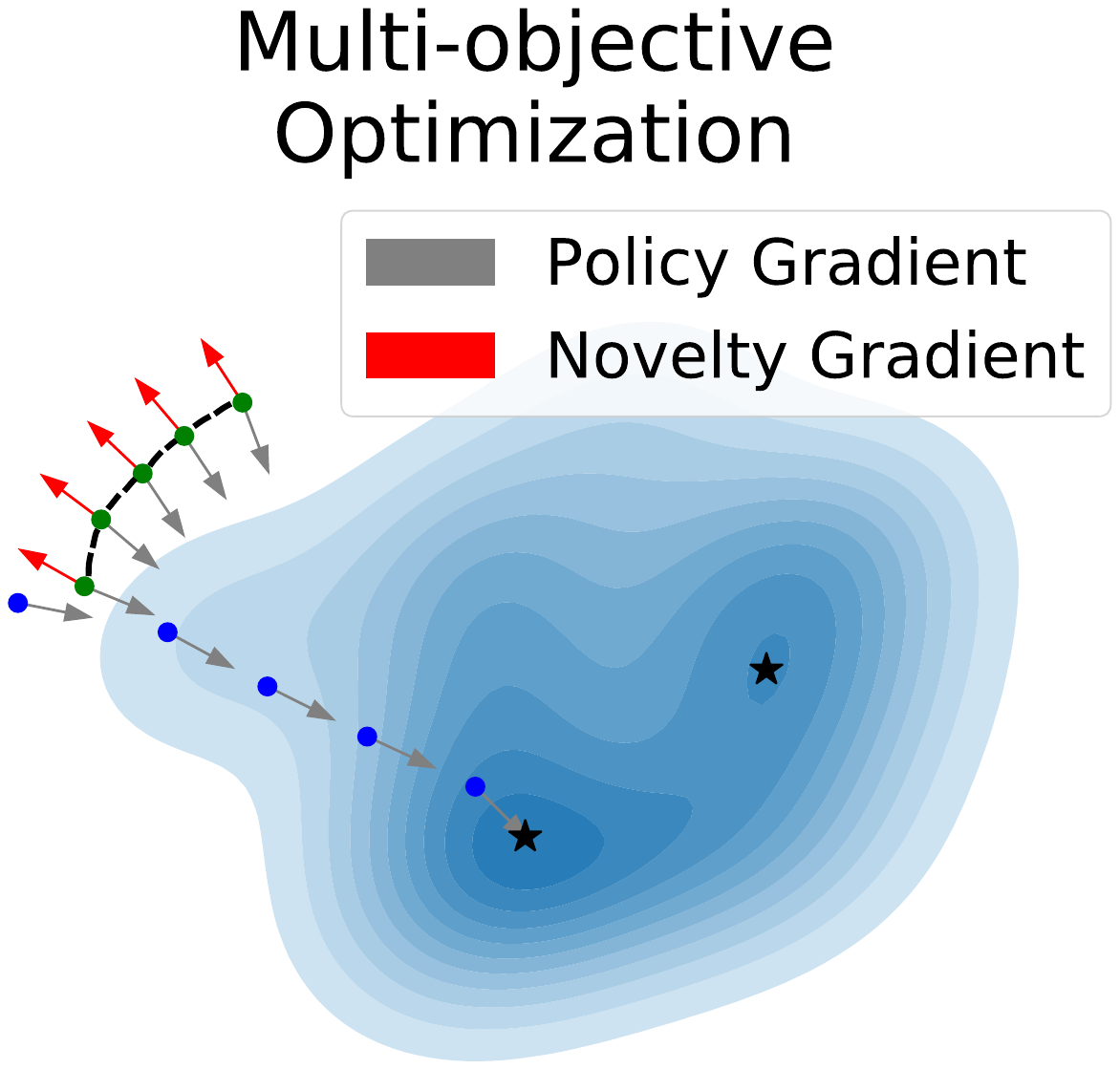}
		\end{minipage}
		\begin{minipage}[htbp]{0.27\linewidth}
			\centering
			\includegraphics[width=0.9\linewidth]{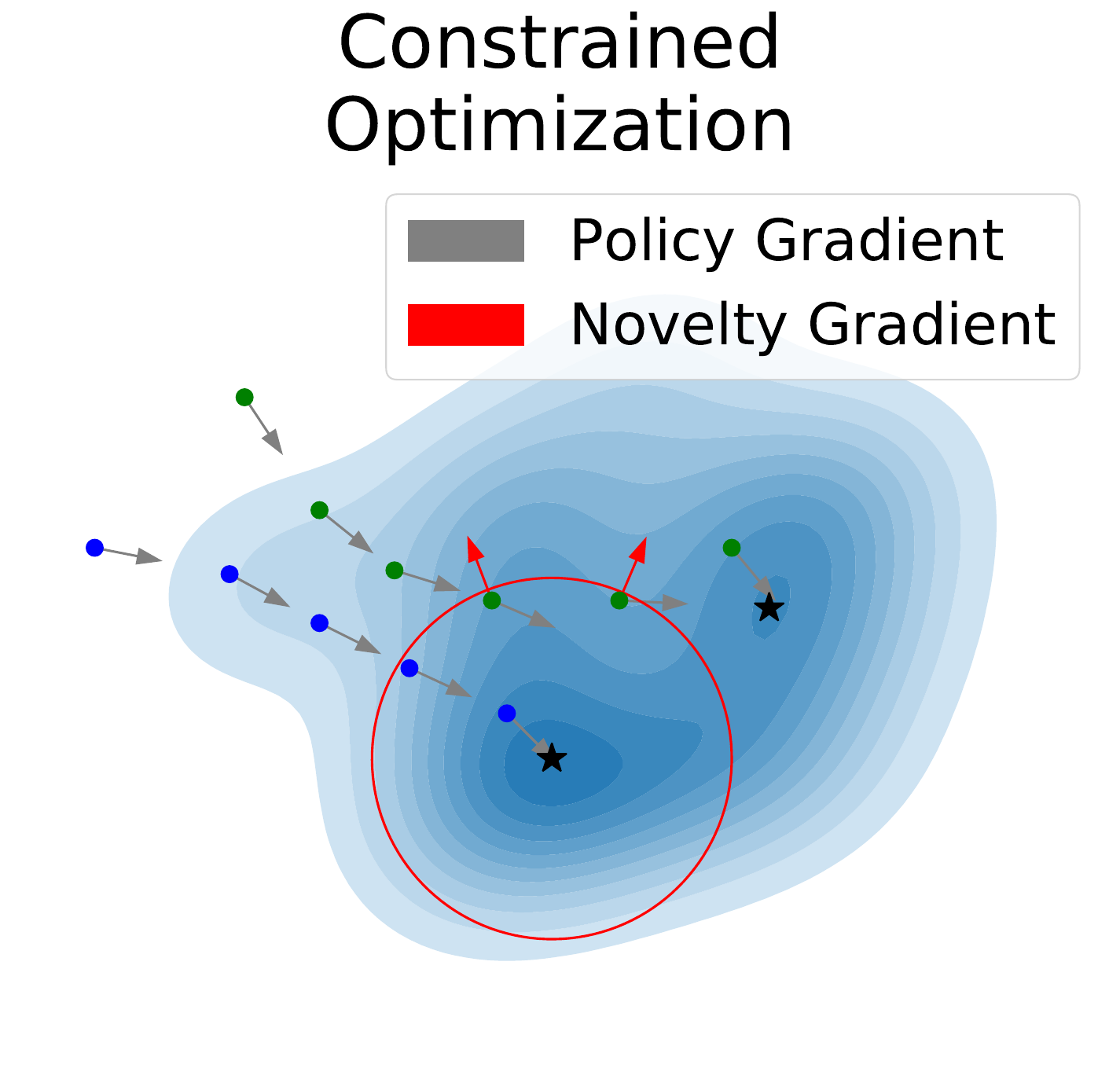}
		\end{minipage}
\vskip -0.2in
\caption{The comparison of the standard policy gradient method without novelty seeking (left), multi-objective optimization method (middle), and our constrained optimization approach (right) for generating novel policies. The standard policy gradient method does not try actively to find novel solutions. The multi-objective optimization method may impede the learning procedure when the novelty gradient is being applied all the time~\citep{DBLP:journals/corr/abs-1905-05252}, e.g., a random initialized policy will be penalized from getting closer to the previous policy due to the conflict of gradients, which limits the learning efficiency and the final performance. On the contrary, the novelty gradient of our constrained optimization approach will only be considered within a certain region to keep the policy being optimized away from highly similar solutions. Such an approach is more flexible and includes the multi-objective optimization method as its special case.}
\label{fig_illu}
\end{figure*}

In previous attempts for novel policy generation, there are three main challenges: (1) heuristically defined metric for novelty estimation is computational expensive~\citep{zhang2019learning}, (2) defining novelty reward for an entire episode yields additional challenge in credit assignment, and (3) solving the problem under the formulation of multi-objective optimization leads to the performance decay in the original task. Fig.~\ref{fig_illu} compares the policy gradients of three cases, namely the one without novel policy seeking, novelty-seeking with multi-objective optimization, and novelty-seeking with constrained optimization methods, respectively. In this work we take into consideration not only the novelty of a set of learned policies but also the performance of those novel policies in the primal task, when addressing the problem of novel-policy-generation. 

\textbf{Our contributions} can be summarized as follows:

\textbf{1.} Mathematically, we introduce a lightweight metric to compute the difference between policies with \textit{instant feedback} at every timestep, to address the first two drawbacks of previous novel policy seeking methods discussed above;

\textbf{2.} Practically, we propose a constrained optimization formulation for novel policy generation to avoid hindering the primal task performance while seeking cross-policy diversity.
We further design an efficient \textit{Novelty-Reward-Scale-Agnostic} algorithm dubbed as IPD, resembling the interior point method in constrained optimization literature; 

\textbf{3.} Empirically, we evaluate IPD on several continuous control benchmarks to generate groups of diverse policies, showing the strengths of our constrained optimization solution for novelty-seeking can generate a series of diverse and well-performing policies, compared to previous multi-objective novel policy generation methods.

\section{Related Work}
\label{sec:relwork}

\textbf{Intrinsic motivation methods.}
\label{privous_work:intrinsic_motivation}
In previous work, different approaches are proposed to provide intrinsic motivation or intrinsic reward as a supplementary to the primal task reward for better exploration~\citep{houthooft2016variational,pathak2017curiosity,burda2018large,burda2018exploration,DBLP:journals/corr/abs-1902-00528}.
All those approaches use the weighted sum of two rewards, the primal rewards provided by environments and the intrinsic rewards provided by different heuristics. On the other hand, the work of DIAYN and DADS~\citep{eysenbach2018diversity,sharma2019dynamics} learn diverse skills without extrinsic reward. Those approaches focus on decomposing diverse skills of a single policy, while our work focuses on learning diverse behaviors among a batch of policies for the same task.

\textbf{Diverse policy generation methods.}
The work of Such et al. shows that different RL algorithms may converge to different policies for the same task~\citep{such2018atari}. 
On the contrary, we are interested in learning different policies through a single algorithm with the capability of avoiding local optimum. 
The work of Pugh et al. establishes a standard framework for understanding and comparing different approaches to search for quality diversity (QD)~\citep{pugh2016quality}. Conti et al. proposes a solution which avoids local optima as well as achieves higher performance by adding novelty search and QD to evolution strategies~\citep{conti2018improving}.
The Task-Novelty Bisector (TNB) \citep{zhang2019learning} aims to solve novel policy generation problem by jointly optimize the extrinsic rewards and novelty rewards defined by an auto-encoder. In this work, we first adopt TNB in the constrained optimization framework, resulting in Contrained TNB, to demonstrate the dilemma between the task performance and novelty pursuance.

\textbf{Constrained Markov Decision Process.}
The Constrained Markov Decision Process (CMDP)~\citep{altman1999constrained} considers the situation where an agent interacts with the environment under certain constraints. Formally, the CMDP can be defined as a tuple $(\mathcal{S},\mathcal{A},\gamma,r,c,C,P,s_0)$, where $\mathcal{S}$ and $\mathcal{A}$ are the state and action space; $\gamma\in[0,1)$ is a discount factor; $r:\mathcal{S}\times\mathcal{A}\times\mathcal{S}\to \mathbb{R}$ and $c: \mathcal{S}\times\mathcal{A}\times\mathcal{S}\to \mathbb{R}$ denote the reward function and cost function; $C\in \mathbb{R}^+$ is the upper bound of permitted expected cumulative cost; $P(\cdot|s,a):\mathcal{S}\times\mathcal{A}\to \mathcal{S}$ denotes the transition dynamics, and $s_0$ is the initial state. Denote the Markovian policy class as $\Pi$, where $\Pi = \{\pi: \mathcal{S}\times\mathcal{A}\to[0,1],\sum_a\pi(a|\pi) = 1\}$
The learning objective of a policy for CMDP is to find a $\pi^*\in\Pi$, such that 
\begin{equation}
\label{cmdp}
\begin{split}
    \pi^*=\max_{\pi\in\Pi} \mathbb{E}_{\tau\sim\pi,s'\sim P}[\sum_{t=0}^\infty \gamma^t r(s,a,s')], ~~ \text{s.t.} ~~\mathbb{E}_{\tau\sim\pi,s'\sim P}[\sum_{t=0}^\infty \gamma^t c(s,a,s')] \le C,
\end{split}
\end{equation}
where $\tau$ indicates a trajectory $(s_0,a_0,s_1,...)$ and $\tau \sim \pi$ represents the distribution over trajectories following policy $\pi$: $a_t\sim  \pi(\cdot|s_t), s_{t+1}\sim P(\cdot|s_t,a_t); t=0,1,2,...$. Previous literature provide several approaches to solve CMDP~\citep{achiam2017constrained,chow2018lyapunov,raybenchmarking,sun2021safe}.

\section{Methods}
\label{sec:frmwork}
In Sec.\ref{method_noveltymetric}, we define a metric space that measures the difference between policies,
which is the fundamental element for the proposed methods. In Sec.\ref{sec_estimation}, we develop a practical estimation method for this metric.
Sec.\ref{sec_formulation} describes the formulation of constrained optimization on novel policy generation. The implementations of two practical algorithms are further introduced in Sec.\ref{sec_practical}.

We denote the policies as $\{\pi_{\theta_i};\theta_i \in \Theta, i=1,2,...\}$, wherein $\theta_i$ represents parameters of the $i$-th policy, $\Theta$ denotes the whole parameter space. In this work, we focus on improving the behavioral diversity of policies from PPO~\citep{schulman2017proximal}, thus we use $\Theta$ to represent $\Theta_\mathrm{PPO}$ in this paper. It is worth noting that the proposed methods can be easily extended to other RL algorithms~\citep{schulman2015trust,lillicrap2015continuous,fujimoto2018addressing,haarnoja2018soft}. To simplify the notation, we omit $\pi$ and denote a policy $\pi_{\theta_i}$ as $\theta_i$ unless stated otherwise.

\subsection{Measuring the Difference between Policies}
\label{method_noveltymetric}


In this work, we use the Wasserstein metric $W_p$~\citep{ruschendorf1985wasserstein,villani2008optimal,pmlr-v70-arjovsky17a} to measure the distance between policies. Concretely, in this work we consider the Gaussian-parameterized policies, where the $W_p$ over two policies can be written in the closed form $W^2_2(\mathcal{N}(m_1,\Sigma_1),\mathcal{N}(m_2,\Sigma_2)) = ||m_1 - m_2||^2 + \text{tr}[\Sigma_1 + \Sigma_2 - 2(\Sigma_1^{1/2}\Sigma_2\Sigma_1^{1/2})^{1/2}] $ as $p=2$, where $m_1,\Sigma_1, m_2, \Sigma_2$ are mean and covariance metrics of the two normal distributions. In the following of this paper, we use $D_W$ to denote the $W_2$ and it is worth noting that when the covariance matrics are identical, the trace term disappears and only the term involving the means remains, i.e., $D_W = |m_1 - m_2|$ for Dirac delta distributions located at points $m_1$ and $m_2$. This diversity metric satisfies the three properties of a metric, namely identity, symmetry as well as triangle inequality.

\begin{proposition}[Metric Space $(\Theta,\overline{D}_W^{q})$]
\label{theo1} 
The expectation of $D_W(\cdot,\cdot)$ of two policies over any state distribution $q(s)$:  
\begin{equation}
\label{eq_prop_1}
\overline{D}_{W}^{q}({\theta_i},{\theta_j}) :={\mathbb{E}}_{s\sim q(s)}[ D_{W}(\theta_i(a|s),\theta_j(a|s))],
\end{equation}
is a metric on $\Theta$, thus $(\Theta,\overline{D}_{W}^{q})$ is a metric space. 
\end{proposition}
The proof of Proposition 1 is straightforward. It is worth mentioning that Jensen Shannon divergence ${D}_{JS}$ or Total Variance Distance ${D}_{TV}$~\citep{endres2003new,fuglede2004jensen,schulman2015trust} can also be applied as alternative metric spaces, 
we choose $D_W$ in our work for that the Wasserstein metric better preserves the continuity~\citep{pmlr-v70-arjovsky17a}.

On top of the metric space $(\Theta,\overline{D}_{W}^{q})$, 
we can then compute the novelty of a policy as follows.
\begin{definition}[Novelty of Policy]
Given a reference policy set $\Theta_{\text{ref}}$ such that $\Theta_{\text{ref}} = \{\theta^{ref}_i,i=1,2,...\}, \Theta_{\text{ref}}\subset\Theta$, the novelty $\mathrm{U}(\theta|\Theta_{\text{ref}})$ of policy $\theta$ is the minimal difference between $\theta$ and all policies in the reference policy set, i.e.,
\begin{equation}
\label{eq_unique}
    \mathrm{U}(\theta|\Theta_{\text{ref}}) := \min_{\theta_j\in \Theta_{\text{ref}}}  \overline{D}_{W}^{q}(\theta,{\theta_j}).
\end{equation}
\end{definition}
Consequently, to encourage the discovery of novel policies discovery, typical novel policy generation methods tend to directly maximize the novelty of a new policy, i.e., $\max_\theta \mathrm{U}(\theta|\Theta_{ref})$, where the $\Theta_{ref}$ includes all existing policies.
\subsection{Estimation of $\overline{D}_{W}^{q}(\theta_i,{\theta_j})$ and the Selection of $q(s)$}
\label{sec_estimation}
In practice, the calculation of $\overline{D}_{W}^{q}(\theta_i,{\theta_j})$ is based on Monte Carlo estimation where we need to sample $s$ from $q(s)$. Although in Eq.(\ref{eq_prop_1}) $q(s)$ can be selected simply as a uniform distribution over the state space, there remains two obstacles: first, in a finite state space we can get precise estimation after establishing ergodicity, but problem arises when facing continuous state spaces due to the difficulty of efficiently obtaining enough samples; second, when $s$ is sampled from a uniform distribution $q$, we can only get \emph{sparse} episodic reward instead of \emph{dense} online reward which is more useful in learning. Therefore, we make an approximation here based on importance sampling.

Formally, we denote the domain of $q(s)$ as $\mathcal{S}_q\subset\mathcal{S}$ and assume $q(s)$ to be a uniform distribution over $\mathcal{S}_q$, without loss of generality in later analysis. Notice $\mathcal{S}_q$ is closely related to the algorithm being used in generating trajectories~\citep{henderson2018deep}. As we only care about the reachable regions of a certain algorithm (in this work, PPO), the domain $\mathcal{S}_q$ can be decomposed by $\mathcal{S}_q = \lim_{N\rightarrow\infty}\bigcup_{i=1}^N \mathcal{S}_{\theta_{i}}$, where $\mathcal{S}_{\theta_i}$ denotes all the possible states a policy $\theta_i$ can visit given a starting state distribution.


In order to get online-reward, we estimate Eq.(\ref{eq_prop_1}) with
\begin{equation}
     \overline{D}_{W}^{q}({\theta_i},{\theta_j}) ={\mathbb{E}}_{s\sim q(s)}[ D_{W}(\theta_i(a|s),\theta_j(a|s))] ={\mathbb{E}}_{s\sim \rho_{\theta_i}(s)}[\frac{q(s)}{\rho_{\theta_i}(s)} D_{W}(\theta_i(a|s),\theta_j(a|s))],
\end{equation}
where we use $\rho_\theta(s)$ to denote the stationary state visitation frequency under policy $\theta$, i.e., $\rho_\theta(s) = P(s_0 = s|\theta) + P(s_1 = s|\theta) +  ... + P(s_T = s|\theta)$ in finite horizon problems. We propose to use the averaged stationary visitation frequency as $q(s)$, e.g., for PPO, $q(s) = \overline{\rho}(s) = \mathbb{E}_{\theta \sim \Theta_\mathrm{PPO}}[\rho_\theta(s)]$. Clearly, choosing $q(s) = \overline{\rho}(s)$ will be much better than choosing a uniform distribution as the importance weight will be closer to $1$. Such an importance sampling process requires a necessary condition that $\rho_{\theta_i}(s)$ and $q(s)$ have the same domain, which can be guaranteed by applying a sufficient exploration noise on $\theta$. 

Another difficulty lies in the estimation of $\overline{\rho}(s)$, which is always intractable given a limited number of trajectories. However, during training, $\theta_i$ is a policy to be optimized and $\theta_j\in\Theta_{ref}$ is a fixed reference policy. The error introduced by approximating the importance weight as $1$ will get larger when $\theta_i$ becomes more distinct from normal policies, at least in terms of the state visitation frequency. We may just regard increasing of the approximation error as the discovery of novel policies.

\begin{proposition}[Unbiased Single Trajectory Estimation]
\label{prop_1}
The estimation of $\rho_\theta(s)$ using a single trajectory $\tau$ is unbiased.
\end{proposition}

The Proposition \ref{prop_1} follows the usual trick in RL that uses a single trajectory to estimate the stationary state visitation frequency. Given the definition of novelty and a practically unbiased sampling method, the next step is to develop an efficient learning algorithm.

\subsection{Constrained Optimization Formulation for Novel Policy Generation}
\label{sec_formulation}

In the traditional RL paradigm, maximizing the expectation of cumulative rewards is commonly used as the objective. i.e., $\max_{\theta \in \Theta} \mathbb{E}_{\tau \sim \theta}[g]$, where $g = \sum_{t=0}\gamma^t r_t$ and $\tau\sim\theta$ denotes a trajectory $\tau$ sampled from the policy $\theta$.


To improve the diversity of different agents' behaviors, the learning objective must take both the reward from the primal task and the policy novelty into consideration. Previous approaches \citep{houthooft2016variational,pathak2017curiosity,burda2018large,burda2018exploration,DBLP:journals/corr/abs-1902-00528} often directly use the weighted sum of these two terms as the objective:
\begin{equation}
\max_{\theta \in \Theta}\mathbb{E}_{\tau \sim {\theta}} {[g_\mathrm{total}]} = 
	\max_{\theta \in \Theta} \mathbb{E}_{\tau \sim {\theta}} {[\alpha \cdot g_\mathrm{task} + (1- \alpha) \cdot g_\mathrm{int}]},
\label{multiobj}
\end{equation}
where $0< \alpha< 1$ is a weight hyper-parameter, $g_\mathrm{task}$ is the reward from the primary task,
and $g_\mathrm{int} = \sum_{t=0} \gamma^t r_{\mathrm{int},t}$ is the cumulative intrinsic reward of the \textit{intrinsic reward} $r_{\mathrm{int},t}$. 
In our case, the intrinsic reward is the novelty reward $r_\mathrm{int} = \min_{\theta_j\in \Theta_{ref}}  \overline{D}_{W}^{\overline{\rho}}(\theta,{\theta_j})$. 
These methods can be summarized as Weighted Sum Reward (WSR) methods~\citep{DBLP:journals/corr/abs-1905-05252}.
Such an objective is sensitive to the selection of $\alpha$ as well as the formulation of $r_\mathrm{int}$. 
For example, in our case formulating the novelty reward $r_\mathrm{int}$ as $\min_{\theta_j} \overline{D}_{W}^{\overline{\rho}}(\theta,{\theta_j})$, 
$\exp{[ \min_{\theta_j}  \overline{D}_{W}^{\overline{\rho}}(\theta,{\theta_j})]}$ 
and $-\exp{[- \min_{\theta_j} \overline{D}_{W}^{\overline{\rho}}(\theta,{\theta_j})]}$ 
will lead to significantly different results as they determine the trade-offs in the two terms given $\alpha$.
Besides, dilemma also arises in the selection of $\alpha$: while a large $\alpha$ may undermine the contribution of intrinsic reward, a small $\alpha$ could ignore the importance of the primal task, leading to the failure of an agent in solving the task. 

The crux of tackling such an issue is to deal with the conflict between different objectives. The work of Zhang et al. proposes the TNB, where the task reward is regarded as the dominant one while the novelty reward is regarded as subordinate~\cite{DBLP:journals/corr/abs-1905-05252}. However, as TNB considers the novelty gradient all the time, it may hinder the learning process. Intuitively, well-performing policies should be more similar to each other than to random initialized policies. As a new random initialized policy is different enough from previous policies, considering the novelty gradient at beginning of training will result in a much slower learning process. 




In order to tackle the above problems and adjust the extent of novelty in new policies, we propose to solve the novelty-seeking problem under the perspective of constrained optimization. The intuition is as follows: while the task reward is considered as a learning objective, the novelty reward should be considered as a bonus instead of another objective, thus should not impede the learning of the primal task. Fig.~\ref{fig_illu} illustrates how novelty gradients impede the learning of a policy: at the beginning of learning, a random initialized policy should learn to be more similar to a well-performing policy rather than be different. The seeking of novelty should not be taken into consideration all the time during learning.
With such an insight, 
we update the multi-objective optimization problem in Eq.(\ref{multiobj}) into a constrained optimization problem as:
\begin{equation}
\begin{array}{cc}
 \max_{\theta\in \Theta}f(\theta) =\mathbb{E}_{\tau \sim {\theta}} {[g_\mathrm{task}]}
   ~~~~ \mathrm{s.t.} ~ g_t(\theta) = \overline{r}_{\mathrm{int},t} - r_0 \geq 0,\forall t=1,2,...,T,
\end{array}
\label{constr}
\end{equation}
where $r_0$ is a threshold indicating minimal permitted novelty, and $\overline{r}_{\mathrm{int},t}$ denotes a moving average of ${r}_{\mathrm{int},t}$. as we need not force every single action of a new agent to be different from others. Instead, we care more about the long-term differences. Therefore, we use cumulative novelty terms as constraints. Moreover, the constraints can be flexibly applied after the first $t_S$ timesteps (e.g.,~ $t_S = 20$) for the consideration of similar starting sequences, so that the constraints can be written as 
$g_t(\theta) \geq 0, \forall t=t_S, ..., T$.

\begin{figure*}[t]
\centering
\hfill
\begin{minipage}{0.35\linewidth}
\centering
\begin{algorithm}[H]
\begin{algorithmic}[1]
\caption{IPD}
\label{algo_IPD}
\small
\STATE {\bfseries Input:} (1) a behavior policy ${\theta_{{old}}}$, (2) a set of previous policies $\{\theta_j\}_{j=1}^M$, (3) a novelty metric $U(\theta,\{\theta_j\}|\rho) = U(\theta,\{\theta_j\}|\tau)= \min_{\theta_j}  \overline{D}_{W}^{\tau}(\theta,{\theta_j})$, (4) a novelty threshold $r_0$ and starting point $t_S$.
\STATE Initialize ${\theta_{{old}}}$.
\FOR{iteration $= 1,2,...$}
    \FOR{t $= 1,2,..., T$}%
        \STATE Step the environment by taking action $a_t\sim \theta_{\textrm{old}}$ and collect transitions.%
        \textcolor{blue}{%
        \IF{$U({\theta_{{old}}},\{\theta_j\}|\tau) - r_0 <0 $ AND $t>t_S$}
            \STATE Break this episode.%
        \ENDIF
        }
    \ENDFOR
    \STATE Update policy parameters based on sampled data.%
\ENDFOR
\end{algorithmic}
\end{algorithm}
\end{minipage} \hfill
\begin{minipage}{0.59\linewidth}
\centering
\begin{algorithm}[H]
\begin{algorithmic}[1]
\caption{Constrained TNB}
\label{algo_CTNB}
\small
	\STATE {\bfseries Input:}  (1-4) same as IPD, (5) \textcolor{blue}{a value network for cost $V_c$}.
\STATE Initialize ${\theta_{{old}}}$.
\FOR{iteration $= 1,2,...$}
    \FOR{t $= 1,2,..., T$}%
        \STATE Step the environment by taking action $a_t\sim \theta_{\textrm{old}}$ and collect transitions.%
    \ENDFOR
    \STATE Compute advantage of reward $\hat{A_{r,1}},...,\hat{A_{r,T}}$.
	\textcolor{blue}{
	\STATE Compute advantage of cost $\hat{A_{c,1}},...,\hat{A_{c,T}}$}.
	 \STATE Optimize objective for reward $\mathcal{L}_r^{\text{CLIP}}$, with gradient $g_r = \nabla_\theta\mathcal{L}_r^{\text{CLIP}}$.
	\textcolor{blue}{%
    	 \STATE Optimize objective for cost $\mathcal{L}_c^{\text{CLIP}}$, with gradient $g_c = -\nabla_\theta\mathcal{L}_c^{\text{CLIP}}$ 
    }%
        \IF{$U({\theta_{{old}}},\{\theta_j\}|\tau) - r_0 <0 $}
        	\STATE Calculate $\vec{p}$ according to Eq.(\ref{eq_tnb_p}) with $g_r$ and $g_c$
        \ELSE
         \STATE Calculate $\vec{p}$ with $g_r$
        \ENDIF
	 \STATE Update policy parameters
\ENDFOR
\end{algorithmic}
\end{algorithm}
\end{minipage}
\hfill
\vskip -0.2 in
\end{figure*}

\subsection{Practical Novel Policy Generation Methods}
\label{sec_practical}

One thing to note is that WSR and TNB proposed in the prior work~\citep{zhang2019learning} correspond to different approaches in constrained optimization problems, yet some important ingredients are missing. We in this section adopt TNB according to the Feasible Direction Method in constrained optimization and then propose our method of Interior Policy Differentiation (IPD), according to the Interior Point Method in constrained optimization literature. A detailed discussion on WSR is provided in Appendix~\ref{appdx_wsr}.

\paragraph{TNB: Feasible Direction Method}
The Feasible Direction Method (FDM)~\citep{ruszczynski1980feasible,herskovits1998feasible} solves the constrained optimization problem by finding a direction $\vec{p}$ where taking gradient upon will lead to the increment of the objective function as well as constraints satisfaction, i.e., $\nabla_{\theta}f^{\rm{T}} \cdot \vec{p} >0$, if $g > 0$ and $\nabla_{\theta}g^{\rm{T}} \cdot \vec{p} >0$ otherwise.
The TNB proposes to use a revised bisector of gradients $\nabla_{\theta}f$ and $\nabla_{\theta}g$ as $\vec{p}$,
\begin{equation}
\label{eq_tnb_p}
\vec{p}=
\begin{cases}
\nabla_{\theta}f + \frac{|\nabla_{\theta}f|}{|\nabla_{\theta}g|} \nabla_{\theta}g
, \quad\quad\quad \mathrm{if}\cos{(\nabla_{\theta}f,\nabla_{\theta}g)} > 0 \\
\nabla_{\theta}f + \frac{|\nabla_{\theta}f|}{|\nabla_{\theta}g|} \nabla_{\theta}g \cdot \cos{(\nabla_{\theta}f,\nabla_{\theta}g)} 
, \quad \text{otherwise}
\end{cases}
\end{equation}
Clearly, Eq.(\ref{eq_tnb_p}) satisfies the constraints but it is more strict than it as the $\nabla_{\theta}g$ term always exists during the optimization of TNB. Based on TNB, we provide a revised approach, named Constrained Task Novel Bisector (CTNB), which resembles better with FDM. Specifically, when $g>0$, CTNB will not apply $\nabla_{\theta}g$ on $g$.
It is clear that TNB is a special case of CTNB when the novelty threshold $r_0$ is set to infinity. We note that in both TNB and CTNB, the learning stride is fixed to be $\frac{|\nabla_{\theta}f| + |\nabla_{\theta}g|}{2}$ and may lead to problem when $\nabla_{\theta}f\rightarrow0$, where the final optimization result will rely heavily on the selection of $g$, i.e., the shape of $g$ is crucial for the success of this approach. We propose CTNB in our work, as a constrained optimization vairant of TNB, to demonstrate the importance of the constrained optimization perspective in novelty seeking, however, we do not in practice observe such a method achieves satisfactory performance.

\paragraph{IPD: Interior Point Method}


\begin{figure*}[t]
\vskip 0.2in
\begin{minipage}[htbp]{0.195\linewidth}
	\centering
	\includegraphics[width=0.8\linewidth]{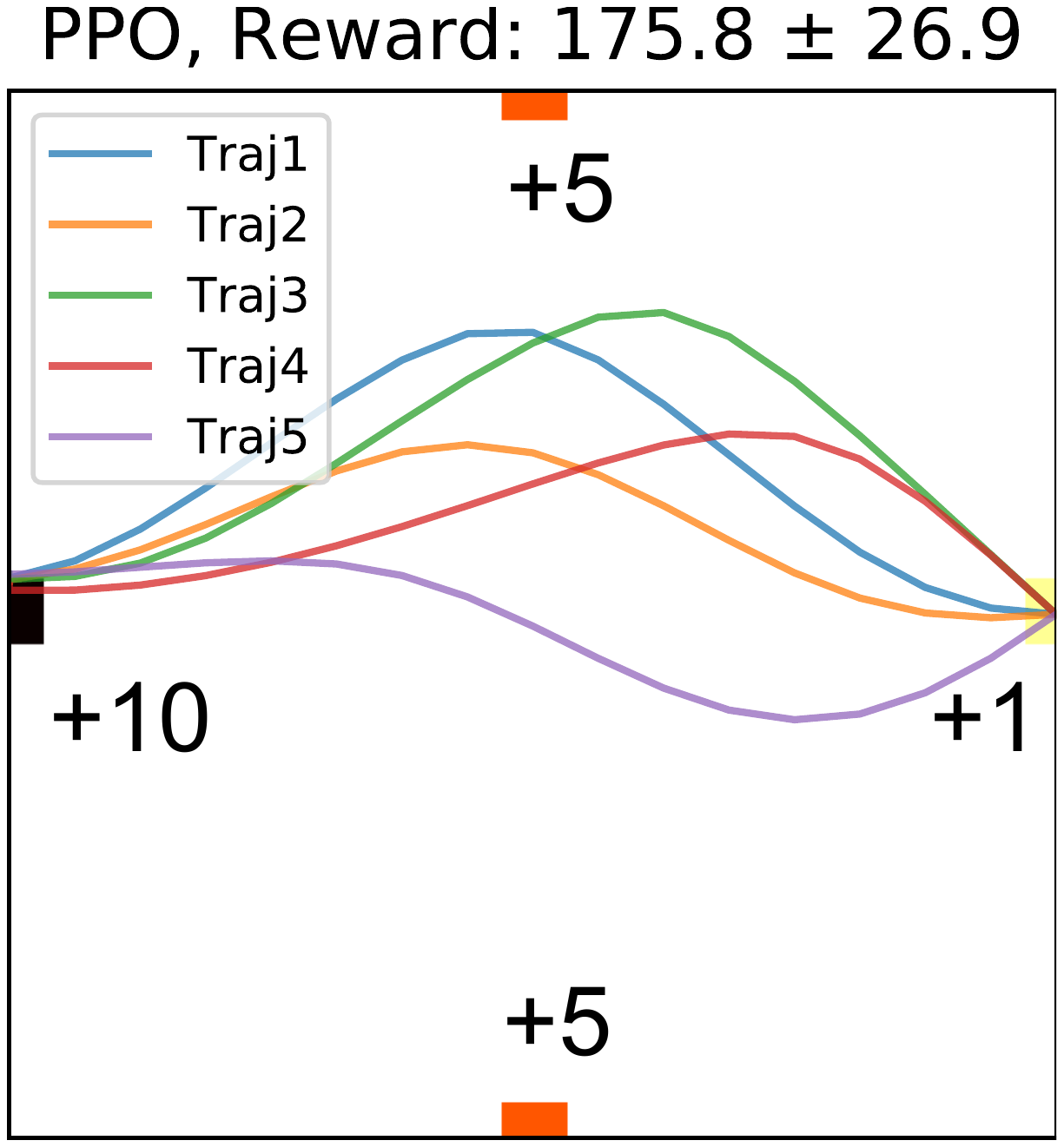}
\end{minipage}%
\begin{minipage}[htbp]{0.195\linewidth}
	\centering
	\includegraphics[width=0.8\linewidth]{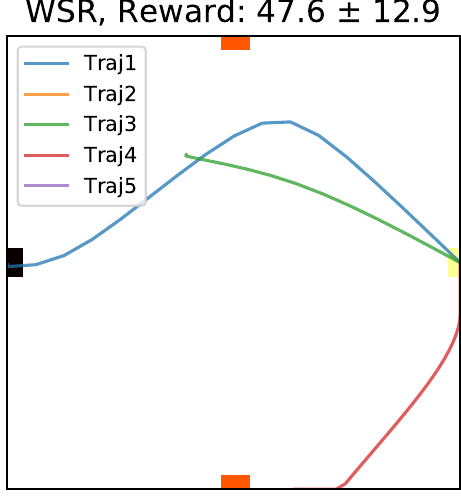}
\end{minipage} %
\begin{minipage}[htbp]{0.195\linewidth}
	\centering
	\includegraphics[width=0.8\linewidth]{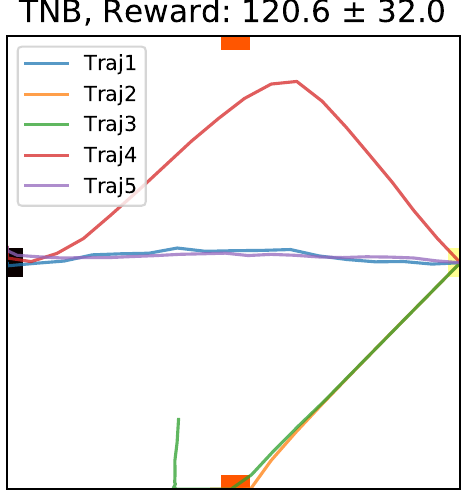}
\end{minipage}%
\begin{minipage}[htbp]{0.195\linewidth}
	\centering
	\includegraphics[width=0.8\linewidth]{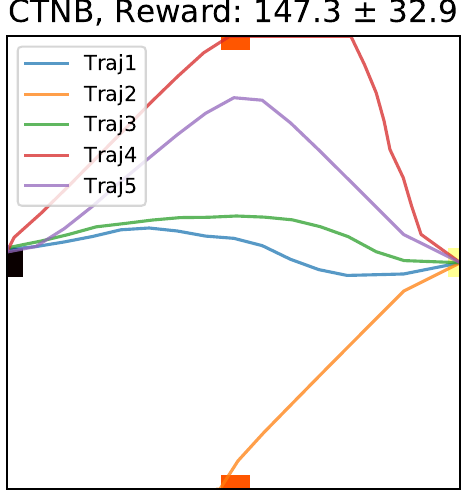}
\end{minipage}%
\begin{minipage}[htbp]{0.195\linewidth}
	\centering
	\includegraphics[width=0.8\linewidth]{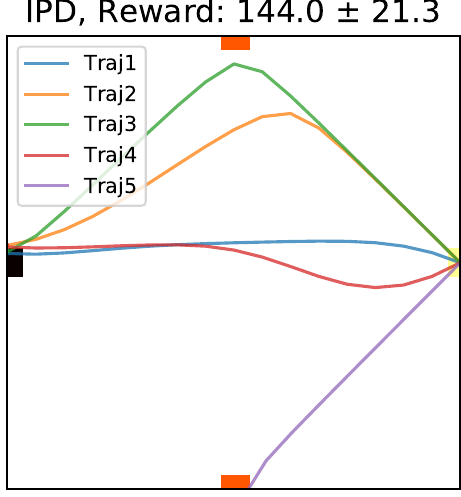}
\end{minipage}%
\caption{Experimental results on the Four Reward Maze Problem. We generate $5$ policies with different novel policy generation methods, and use the PPO with different random seeds as baseline. In each figure, the $5$ lines indicate 5 trajectories when the game is started from the right hand side. It worth noting that the results of WSR, CTNB and IPD are associated with the parameters of weights or threshold. We set the weight parameter in WSR as $10$ to make the two reward terms comparable, and set the thresholds in CTNB and IPD as the averaged novelty between policies trained with PPO. All policies are trained with $6.1\times 10^3$ episodes.}
\label{toy_model_result}
\vskip -0.2in
\end{figure*}


The Interior Point Method~\citep{potra2000interior,dantzig2006linear} is another approach used to solve the constrained optimization problem. Thus here we solve Eq.(\ref{constr}) using the Interior Policy Differentiation (IPD), which can be regarded as an analogy of the Interior Point Method.
In the vanilla Interior Point Method, the constrained optimization problem in Eq.(\ref{constr}) is solved by reforming it to an unconstrained form with an additional barrier term $- \alpha \log{g(\theta)}$ in the objective as
$\max_{\theta\in \Theta} f(\theta) - \alpha \log{g(\theta)}$, 
or more precisely in our problem with the formulation with Eq.(\ref{constr}) we have
$\max_{\theta\in \Theta} \mathbb{E}_{\tau \sim {\theta}}[g_\mathrm{task} - \sum_{t=0}^{T}\alpha\log{(\overline{r}_{\mathrm{int},t} - r_0)}]$, 
where $\alpha>0$ is the barrier factor.
Besides the log barrier term, there are other choices like $\alpha \frac{1}{g(\theta)}$ can be used and the objective becomes $\max_{\theta\in \Theta} \  f(\theta) + \alpha \frac{1}{g(\theta)}$.
As $\alpha$ is small, the barrier term will introduce only minuscule influence on the objective. On the other hand, when $\theta$ get closer to the barrier, the objective will increase rapidly. 
The limits when $\alpha\rightarrow0$ then lead to the solution of Eq.(\ref{constr}). The convergence of such methods are provided in previous works~\citep{conn1997globally,wright2001convergence}. 

However, directly applying IPM is computationally expensive and numerically unstable. In this work, we propose a simple yet novel heuristic method that resembles the idea of barrier methods: we implicitly apply such barrier terms by providing termination signals in interactions with the environments. Our method can be regarded as revising the primal task MDP into a new one in which the behaviors of agents must satisfy novelty constraints.
Specifically, in the RL paradigm, the learning procedure of an agent is determined by the experiences collected during interactions with the environment and the sampling strategy used to filter experiences in the calculation of policy gradients. Since the learning process is based on sampled transitions, a more natural way can thus be used to perform the constrained optimization. We can simply bound the collected transitions in the feasible region by permitting previously trained $M$ policies $\theta_i \in \Theta_{\text{ref}}, i=1,2,...,M$ sending termination signals during the training process of new agents. 
In other words, we implicitly bound the feasible region by terminating any new agent that steps outside it.

Consequently, during the training process, all valid samples we collected are inside the feasible region, which means these samples are less likely to appear in previously trained policies. At the end of the training, we obtain a new policy that has sufficient novelty. In this way, we no longer need to consider the trade-off between intrinsic and extrinsic rewards deliberately. The learning process of IPD is thus more robust and no longer suffers from the objective inconsistency. 
\begin{remark}
[Reward-Shaping-Agnostic Novelty Seeking] IPD is a gradient-free method with regard to the novelty reward.
\end{remark}
Remark 1 is an important property that only IPD owns. For other approaches, including both multi-objective approaches and constrained optimization approaches, an elaborated design of the novelty reward function is needed, e.g., it can be any monotonic increasing function of $\overline{D}^{\overline{\rho}}_W$. While it is well-known that reward shaping~\citep{randlov1998learning,laud2004theory} is in general a non-trivial work that requires domain knowledge~\citep{vinyals2019grandmaster,akkaya2019solving,berner2019dota,elbarbari2021ltlf}, such an additional novelty-reward term further increases the burden of proper reward design.

Differently, the seeking of novelty in IPD does not require any (policy) gradient information that flows from the novelty reward, therefore, the selection of reward scaling function is agnostic to the performance of IPD. IPD learns to become novel in a passive manner, i.e., an episode will be terminated whenever the averaged step-wise novelty is lower than a given threshold. Searching in a constant hyper-parameter space is at least tractable while searching in a monotonically increasing functional reward shaping class~\citep{zhang2019learning} is not -- let alone IPD works well with a default novelty threshold parameter $r_0=$ averaged differences between PPO policies.
\begin{figure*}[t]
\vskip 0.2in
\begin{center}
\includegraphics[width=0.9\linewidth]{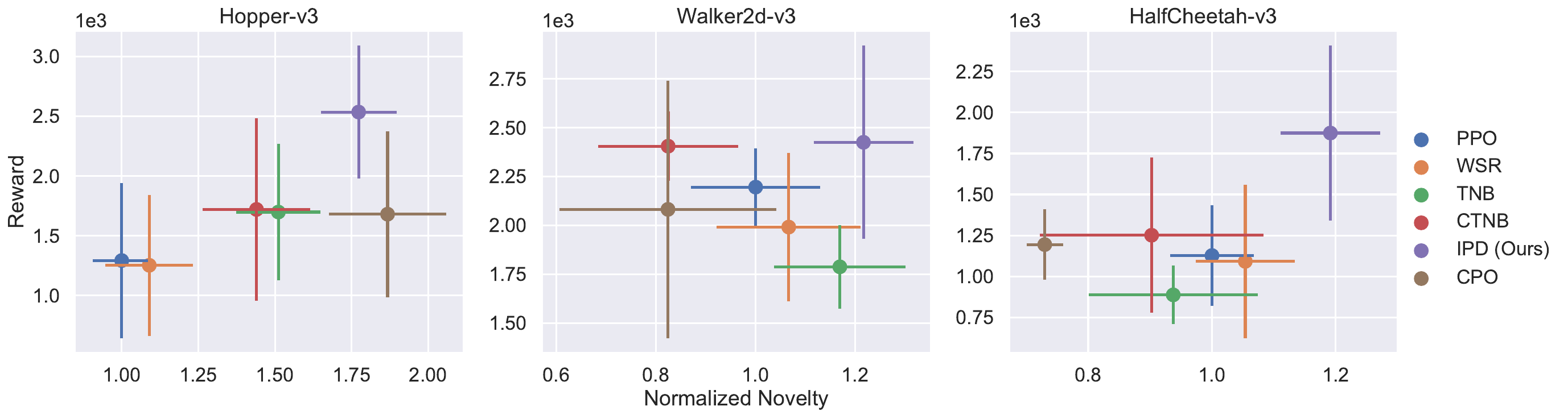}
\caption{The performance and novelty comparison of different methods in Hopper-v3, Walker2d-v3 and HalfCheetah-v3 environments. The value of novelty is normalized to relative novelty by regarding the averaged novelty of PPO policies as the baseline. The results are from $10$ policies of each method, with the points showing their mean and lines showing their standard deviation.}
\label{fig_novelty}
\end{center}
\vskip -0.2in
\end{figure*}

\section{Experiments}
\label{sec_exp}

According to Proposition 2, the novelty reward $r_{int}$ in Eq.(\ref{constr}) under our novelty metric can be unbiasedly approximated by $r_{\mathrm{int}}=\min_{\theta_j\in \Theta_{ref}}  \overline{D}_{W}^{\rho_\theta}(\theta(a|s_t),\theta_j(a_j|s_t))$. We thus utilize this novelty metric directly throughout our experiments. We apply different novel policy generation methods, namely WSR, TNB, CTNB, and IPD, to the backbone RL algorithm PPO~\citep{schulman2017proximal}. 
The extension to other popular RL algorithms is straightforward. More implementation details are depicted in Appendix \ref{ImpleDetail}.

Experiments in the work of \cite{henderson2018deep} show that one can simply change the random seeds before training to get policies that perform differently.
Therefore, we use PPO with varying random seeds as a baseline method for novel policy generation and use the averaged differences between policies learned by this baseline as the \textbf{default threshold} $r_0$ in CTNB and IPD.
Algorithm~\ref{algo_IPD} and Algorithm~\ref{algo_CTNB} show the pseudo code of IPD and CTNB based on PPO, where the blue lines show the additional code added to the standard PPO. Qualitative results can be found in Appendix~\ref{quali_results}.

\begin{table*}
\centering
\caption{Reward and Success Rate of $10$ Policies. IPD beat CTNB, CPO, TNB and WSR in all three environments. Constrained optimization approaches outperforms multi-objective methods.}
\label{tab_performance}
\begin{small}
	\centering
	\begin{tabular}{ccccccc}
		\toprule
		\multicolumn{1}{l}{ }& \multicolumn{3}{c}{Reward} & \multicolumn{3}{c}{Success Rate} \\
		\midrule
		 Environment     & Hopper     & Walker2d  & HalfCheetah  & Hopper     & Walker2d  & HalfCheetah \\
		\midrule
		PPO    & $1292\pm 650$ &$ 2196\pm 200 $ & $1127 \pm 308$ & $ 0.5$ &$ 0.5 $ & $ 0.5$ \cr
	WSR   & $1253\pm 591$&$1992\pm 380$ & $ 1091 \pm 469$ & $ 0.6 $&$ 0.3 $ & $ 0.3$  \cr
	TNB   & $1699\pm 573$& $1788\pm 214$ & $887 \pm 178$ & $0.8 $& $ 0.0 $ & $ 0.1$ \cr
	CPO & $1681\pm 696 $ & $2082\pm 660$ & $1194\pm 215 $ &$0.8$ & $0.6$ & $0.8$ \cr
	CTNB   & $1721\pm 765 $& $\bm{2405 \pm 177}$ & $ 1251  \pm 473$ & $ 0.8$& $ \bm{0.9} $ & $ 0.5$ \cr
    IPD (Ours)   &$\bm{2536\pm557}$ & $2282 \pm 206$ & $\bm{1875\pm 533}$ &$ \bm{1.0}$ & $ 0.6 $ & $ \bm{0.9} $ \cr
		\bottomrule
	\end{tabular}
\end{small}
\end{table*}

\subsection{The Four Reward Maze Problem} 
We first utilize a basic 2-D environment named Four Reward Maze as a diagnostic environment where we can visualize learned policies directly. In this environment, four positive rewards of different values (e.g.,~ $+5, +5, +10, +1$ for top, down, left and right respectively) are assigned to four middle points with radius $1$ on each edge in a 2-D $N\times N$ square map. We use $N=16$ in our experiments. The observation of a policy is the current position and the agent will receive a negative reward of $-0.01$ at each timestep except stepping into the reward regions. Each episode starts from a randomly initialized position and the action space is limited to $[-1,1]$. The performance of each agent is evaluated by the averaged performances over 100 trials.

Results are shown in Fig.~\ref{toy_model_result},
where the behaviors of the PPO agents are quite similar, suggesting the diversity provided by random seeds is limited. WSR and TNB solve the novel policy generation problem from the multi-objective optimization formulation, they thus suffer from the unbalance between performance and novelty.
While WSR and TNB both provide sufficient novelty,
performances of agents learned by WSR decay significantly,
so did TNB due to an encumbered learning process, 
as we analyzed in Sec.\ref{sec_formulation}.
Both CTNB and IPD, solving the task with novelty-seeking from the constrained optimization formulation, provide evident behavior diversity and perform recognizably better than TNB and WSR.

\begin{figure*}[t]
\begin{minipage}[htbp]{0.33\linewidth}
			\centering
			\includegraphics[width=0.97\linewidth]{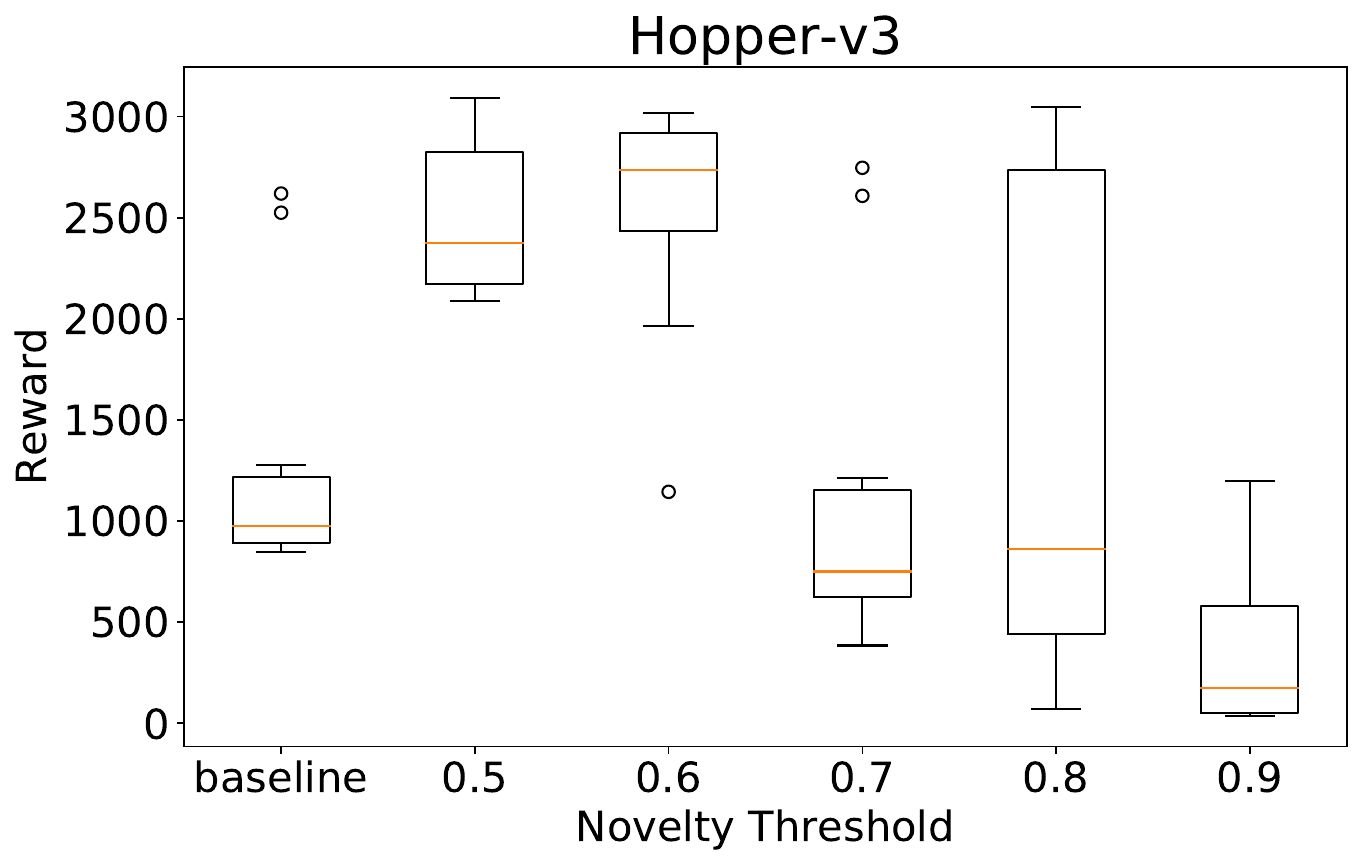}
		\end{minipage}%
		\begin{minipage}[htbp]{0.33\linewidth}
			\centering
			\includegraphics[width=0.97\linewidth]{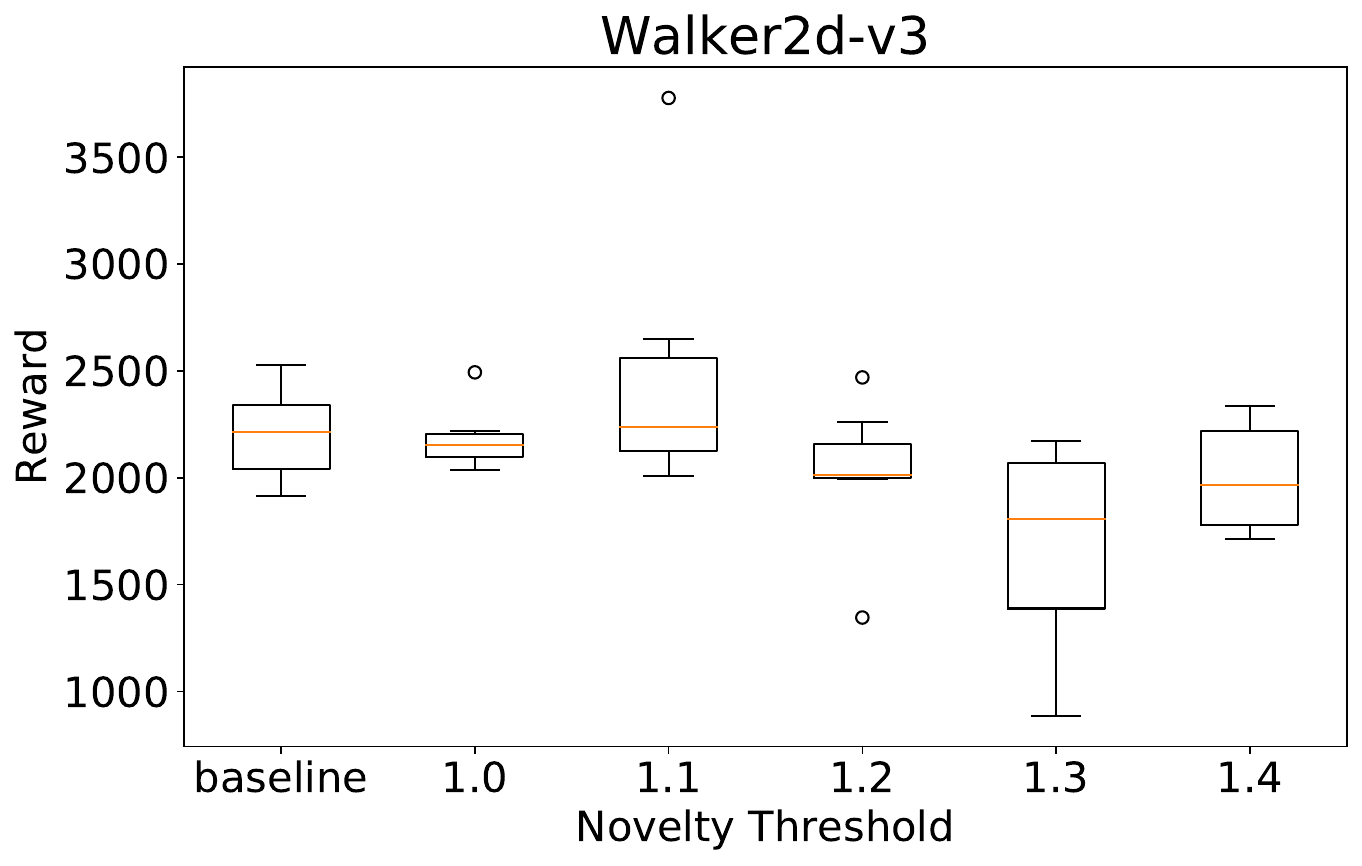}
		\end{minipage}
		\begin{minipage}[htbp]{0.33\linewidth}
			\centering
			\includegraphics[width=0.97\linewidth]{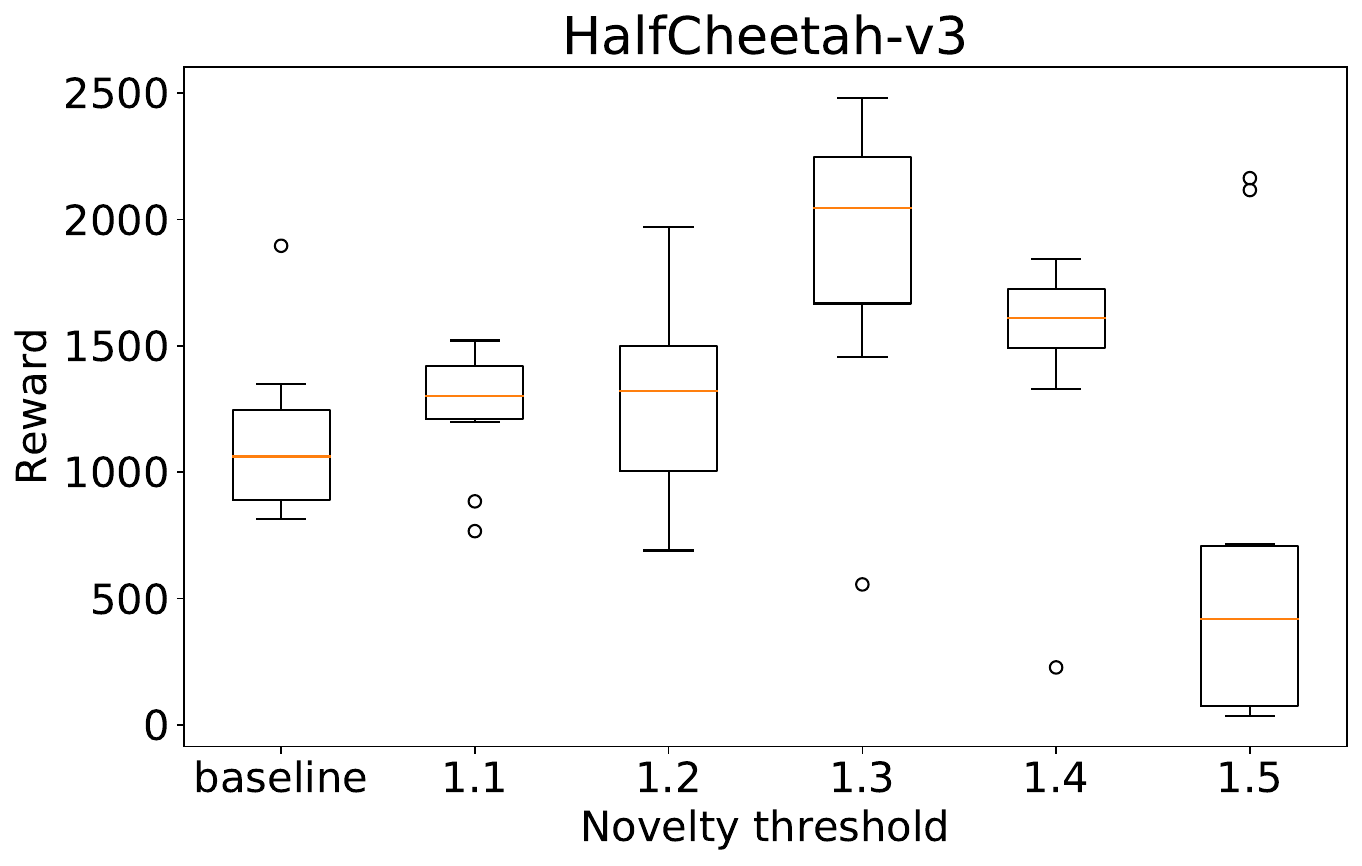}
		\end{minipage}
\caption{The performance under different novelty thresholds in the Hopper, Walker and HalfCheetah environments. The results are collected from 10 learned policies based on PPO. The box extends from the lower to upper quartile values of the data, with a line at the median. The whiskers extend from the box to show the range of the data. Flier points are those past the end of the whiskers.}
\label{fig_ablation}
\vskip -0.2 in
\end{figure*}

\subsection{The MuJoCo Benchmark} 
We evaluate our proposed method on three locomotion tasks~\citep{1606.01540,TodorovET12}: the Hopper-v3 (11 observations and 3 actions), Walker2d-v3 (11 observations and 6 actions), and HalfCheetah-v3 (17 observations and 6 actions). Although relaxing the healthy termination thresholds in Hopper and Walker may permit more visible behavior diversity, all the environment parameters are set as default values in our experiments to demonstrate the generality of our method. 
\paragraph{Comparison on Novelty and Performance}
We implement WSR, TNB, CTNB, and IPD using the same hyper-parameter settings per environment. And we also apply CPO~\citep{achiam2017constrained} as a baseline as a solution of CMDP. For each method, we first train $10$ policies using PPO with different random seeds. Those PPO policies are used as the primal reference policies, and then we train $10$ novel policies that try to be different from previous reference policies.
Concretely, in each method, the $1{st}$ novel policy is trained to be different from the previous $10$ PPO policies, and the $2{nd}$ should be different from the previous $11$ policies, and so on. More implementation details are depicted in Appendix \ref{ImpleDetail}.

Fig.~\ref{fig_novelty} shows our experimental results in terms of novelty (the x-axis) and the performance (the y-axis). Policies close to the upper right corner are the more novel ones with higher performance.
In all environments, the performance of CTNB, IPD and CPO outperforms WSR and TNB, showing the advantage of constrained optimization approaches in novel policy generation. Specifically, the results of CTNB are all better than their multi-objective counterparts from TNB, showing the superiority of generating novel policies with constrained optimization. In all experiments we use a linear novelty reward function, i.e., $r_\mathrm{int}=\min_{\theta_j} \overline{D}_{W}^{\overline{\rho}}(\theta,{\theta_j})$. We attribute the failure of CPO, TNB and CTNB in Walker and HalfCheetah in finding novel policies to that their convergence behavior is fully controlled by the reward scaling function. Whereas in IPD, there is no novelty-gradient controlled by such a scaling function.

Comparisons of the task-related rewards are carried out in Table \ref{tab_performance}, where among all the four methods, IPD provides sufficient diversity with minimum loss of performance. Instead of performance decay, we find IPD is able to find better policies in the environment of Hopper and HalfCheetah. Moreover, in the Hopper environment, while the agents trained with PPO tend to fall into the same local minimum. (e.g.,~they all jump as far as possible and then terminate this episode. On the contrary, PPO with IPD keeps new agents away from falling into the same local minimum, because once an agent has reached some local minimum, agents learned later will try to avoid this region due to the novelty constraints.
Such property shows that IPD can enhance the traditional RL schemes to tackle the local exploration challenge~\citep{tessler2019distributional,ciosek2019better}. 
A similar feature brings about reward growth in the environment of HalfCheetah.

\paragraph{Success Rate of Each Method}
In addition to averaged reward, we also use the success rate as another metric to compare the performance of different approaches. Roughly speaking, the success rate evaluates the stability of each method in terms of generating a policy that performs as good as the policies PPO generates.
In this work, we regard a policy successful when its performance achieves at least as good as the median performance of policies trained with PPO. To be specific, we use the median of the final performance of PPO as the baseline, and if a novel policy, which aims at performing differently to solve the same task, surpasses the baseline during its training process, it will be regarded as a successful policy. By definition, the success rate of PPO is $0.5$ as a baseline for every environment. Table \ref{tab_performance} shows the success rate of all the methods. The results show that all constrained novel policy generation methods (CTNB, IPD, CPO) can surpass the average baseline during training, while the multi-objective optimization approaches normally can not. 

\subsection{Novel Policy Generation without Performance Decay}
Multi-objective formulation of novel policy generation has the risk of sacrificing the primal performance as the overall objective needs to consider both novelty and primal task rewards. On the contrary, under the perspective of constrained optimization, there will be no more trade-off between novelty and final reward as the only objective is the task reward. Given a certain novelty threshold, the algorithms tend to find the optimal solution in terms of task reward under constraints, thus the learning process becomes more controllable and reliable, i.e., one can utilize the novelty threshold to control the degree of novelty.
The proper magnitude of the novelty threshold leads to more exploration among a population of policies, thus the performance of latterly found policies may be better than or at least as good as those trained without novelty seeking. However, when a larger magnitude of novelty threshold is applied, the performance of found novel policies will decrease because finding a feasible solution will get harder under more strict constraints. Fig.~\ref{fig_ablation} shows 
experimental results on adjusting the thresholds, which supports our intuition.

\section{Conclusion}
In this work, we rethink the novel policy seeking problem under the perspective of constrained optimization. We first introduce a new metric to measure the distances between policies, on top of which we define the novelty of a policy. Based on our formulation of constrained optimization, we provide practical algorithms for constrained novel policy learning, we evaluate several constrained policy optimization methods: namely the CPO, Constrained TNB, and the Interior Policy Differentiation (IPD) proposed in this work. 
Our experimental results demonstrate IPD, as a novelty (policy) gradient-free approach, can effectively learn various well-performing yet diverse policies, outperforming previous multi-objective methods, as well as constrained optimization baselines.

\bibliography{icml2020}

\begin{thebibliography}{8}
\providecommand{\natexlab}[1]{#1}
\providecommand{\url}[1]{\texttt{#1}}
\expandafter\ifx\csname urlstyle\endcsname\relax
  \providecommand{\doi}[1]{doi: #1}\else
  \providecommand{\doi}{doi: \begingroup \urlstyle{rm}\Url}\fi

\bibitem[Author(2021)]{anonymous}
Author, N.~N.
\newblock Suppressed for anonymity, 2021.

\bibitem[Duda et~al.(2000)Duda, Hart, and Stork]{DudaHart2nd}
Duda, R.~O., Hart, P.~E., and Stork, D.~G.
\newblock \emph{Pattern Classification}.
\newblock John Wiley and Sons, 2nd edition, 2000.

\bibitem[Kearns(1989)]{kearns89}
Kearns, M.~J.
\newblock \emph{Computational Complexity of Machine Learning}.
\newblock PhD thesis, Department of Computer Science, Harvard University, 1989.

\bibitem[Langley(2000)]{langley00}
Langley, P.
\newblock Crafting papers on machine learning.
\newblock In Langley, P. (ed.), \emph{Proceedings of the 17th International
  Conference on Machine Learning (ICML 2000)}, pp.\  1207--1216, Stanford, CA,
  2000. Morgan Kaufmann.

\bibitem[Michalski et~al.(1983)Michalski, Carbonell, and
  Mitchell]{MachineLearningI}
Michalski, R.~S., Carbonell, J.~G., and Mitchell, T.~M. (eds.).
\newblock \emph{Machine Learning: An Artificial Intelligence Approach, Vol. I}.
\newblock Tioga, Palo Alto, CA, 1983.

\bibitem[Mitchell(1980)]{mitchell80}
Mitchell, T.~M.
\newblock The need for biases in learning generalizations.
\newblock Technical report, Computer Science Department, Rutgers University,
  New Brunswick, MA, 1980.

\bibitem[Newell \& Rosenbloom(1981)Newell and Rosenbloom]{Newell81}
Newell, A. and Rosenbloom, P.~S.
\newblock Mechanisms of skill acquisition and the law of practice.
\newblock In Anderson, J.~R. (ed.), \emph{Cognitive Skills and Their
  Acquisition}, chapter~1, pp.\  1--51. Lawrence Erlbaum Associates, Inc.,
  Hillsdale, NJ, 1981.

\bibitem[Samuel(1959)]{Samuel59}
Samuel, A.~L.
\newblock Some studies in machine learning using the game of checkers.
\newblock \emph{IBM Journal of Research and Development}, 3\penalty0
  (3):\penalty0 211--229, 1959.

\end{thebibliography}


\begin{thebibliography}{10}

\bibitem{sutton1998introduction}
R.~S. Sutton, A.~G. Barto, {\em et~al.}, {\em Introduction to reinforcement
  learning}, vol.~2.
\newblock MIT press Cambridge, 1998.

\bibitem{vinyals2019grandmaster}
O.~Vinyals, I.~Babuschkin, W.~M. Czarnecki, M.~Mathieu, A.~Dudzik, J.~Chung,
  D.~H. Choi, R.~Powell, T.~Ewalds, P.~Georgiev, {\em et~al.}, ``Grandmaster
  level in starcraft ii using multi-agent reinforcement learning,'' {\em
  Nature}, vol.~575, no.~7782, pp.~350--354, 2019.

\bibitem{akkaya2019solving}
I.~Akkaya, M.~Andrychowicz, M.~Chociej, M.~Litwin, B.~McGrew, A.~Petron,
  A.~Paino, M.~Plappert, G.~Powell, R.~Ribas, {\em et~al.}, ``Solving rubik's
  cube with a robot hand,'' {\em arXiv preprint arXiv:1910.07113}, 2019.

\bibitem{berner2019dota}
C.~Berner, G.~Brockman, B.~Chan, V.~Cheung, P.~D{e}biak, C.~Dennison, D.~Farhi,
  Q.~Fischer, S.~Hashme, C.~Hesse, {\em et~al.}, ``Dota 2 with large scale deep
  reinforcement learning,'' {\em arXiv preprint arXiv:1912.06680}, 2019.

\bibitem{badia2020agent57}
A.~P. Badia, B.~Piot, S.~Kapturowski, P.~Sprechmann, A.~Vitvitskyi, Z.~D. Guo,
  and C.~Blundell, ``Agent57: Outperforming the atari human benchmark,'' in
  {\em International Conference on Machine Learning}, pp.~507--517, PMLR, 2020.

\bibitem{elbarbari2021ltlf}
M.~Elbarbari, K.~Efthymiadis, B.~Vanderborght, and A.~Now{\'e}, ``Ltlf-based
  reward shaping for reinforcement learning,'' in {\em Adaptive and Learning
  Agents Workshop 2021}, 2021.

\bibitem{rogoff1990apprenticeship}
B.~Rogoff, {\em Apprenticeship in thinking: Cognitive development in social
  context.}
\newblock Oxford university press, 1990.

\bibitem{ryan2000intrinsic}
R.~M. Ryan and E.~L. Deci, ``Intrinsic and extrinsic motivations: Classic
  definitions and new directions,'' {\em Contemporary educational psychology},
  vol.~25, no.~1, pp.~54--67, 2000.

\bibitem{van2011social}
C.~P. van Schaik and J.~M. Burkart, ``Social learning and evolution: the
  cultural intelligence hypothesis,'' {\em Philosophical Transactions of the
  Royal Society B: Biological Sciences}, vol.~366, no.~1567, pp.~1008--1016,
  2011.

\bibitem{henrich2017secret}
J.~Henrich, {\em The secret of our success: How culture is driving human
  evolution, domesticating our species, and making us smarter}.
\newblock Princeton University Press, 2017.

\bibitem{harari2014sapiens}
Y.~N. Harari, {\em Sapiens: A brief history of humankind}.
\newblock Random House, 2014.

\bibitem{jaques2019social}
N.~Jaques, A.~Lazaridou, E.~Hughes, C.~Gulcehre, P.~Ortega, D.~Strouse, J.~Z.
  Leibo, and N.~De~Freitas, ``Social influence as intrinsic motivation for
  multi-agent deep reinforcement learning,'' in {\em International Conference
  on Machine Learning}, pp.~3040--3049, 2019.

\bibitem{hughes2018inequity}
E.~Hughes, J.~Z. Leibo, M.~Phillips, K.~Tuyls, E.~Due{\~n}ez-Guzman, A.~G.
  Casta{\~n}eda, I.~Dunning, T.~Zhu, K.~McKee, R.~Koster, {\em et~al.},
  ``Inequity aversion improves cooperation in intertemporal social dilemmas,''
  in {\em Advances in neural information processing systems}, pp.~3326--3336,
  2018.

\bibitem{sequeira2011emerging}
P.~Sequeira, F.~S. Melo, R.~Prada, and A.~Paiva, ``Emerging social awareness:
  Exploring intrinsic motivation in multiagent learning,'' in {\em 2011 IEEE
  International Conference on Development and Learning (ICDL)}, vol.~2,
  pp.~1--6, IEEE, 2011.

\bibitem{peysakhovich2017consequentialist}
A.~Peysakhovich and A.~Lerer, ``Consequentialist conditional cooperation in
  social dilemmas with imperfect information,'' {\em arXiv preprint
  arXiv:1710.06975}, 2017.

\bibitem{zhang2019learning}
Y.~Zhang, W.~Yu, and G.~Turk, ``Learning novel policies for tasks,'' in {\em
  International Conference on Machine Learning}, pp.~7483--7492, PMLR, 2019.

\bibitem{DBLP:journals/corr/abs-1905-05252}
Y.~Zhang, W.~Yu, and G.~Turk, ``Learning novel policies for tasks,'' {\em
  CoRR}, vol.~abs/1905.05252, 2019.

\bibitem{houthooft2016variational}
R.~Houthooft, X.~Chen, Y.~Duan, J.~Schulman, F.~De~Turck, and P.~Abbeel,
  ``Variational information maximizing exploration,'' 2016.

\bibitem{pathak2017curiosity}
D.~Pathak, P.~Agrawal, A.~A. Efros, and T.~Darrell, ``Curiosity-driven
  exploration by self-supervised prediction,'' in {\em Proceedings of the IEEE
  Conference on Computer Vision and Pattern Recognition Workshops}, pp.~16--17,
  2017.

\bibitem{burda2018large}
Y.~Burda, H.~Edwards, D.~Pathak, A.~Storkey, T.~Darrell, and A.~A. Efros,
  ``Large-scale study of curiosity-driven learning,'' {\em arXiv preprint
  arXiv:1808.04355}, 2018.

\bibitem{burda2018exploration}
Y.~Burda, H.~Edwards, A.~Storkey, and O.~Klimov, ``Exploration by random
  network distillation,'' {\em arXiv preprint arXiv:1810.12894}, 2018.

\bibitem{DBLP:journals/corr/abs-1902-00528}
H.~Liu, A.~Trott, R.~Socher, and C.~Xiong, ``Competitive experience replay,''
  {\em CoRR}, vol.~abs/1902.00528, 2019.

\bibitem{eysenbach2018diversity}
B.~Eysenbach, A.~Gupta, J.~Ibarz, and S.~Levine, ``Diversity is all you need:
  Learning skills without a reward function,'' {\em arXiv preprint
  arXiv:1802.06070}, 2018.

\bibitem{sharma2019dynamics}
A.~Sharma, S.~Gu, S.~Levine, V.~Kumar, and K.~Hausman, ``Dynamics-aware
  unsupervised discovery of skills,'' {\em arXiv preprint arXiv:1907.01657},
  2019.

\bibitem{such2018atari}
F.~P. Such, V.~Madhavan, R.~Liu, R.~Wang, P.~S. Castro, Y.~Li, L.~Schubert,
  M.~Bellemare, J.~Clune, and J.~Lehman, ``An atari model zoo for analyzing,
  visualizing, and comparing deep reinforcement learning agents,'' {\em arXiv
  preprint arXiv:1812.07069}, 2018.

\bibitem{pugh2016quality}
J.~K. Pugh, L.~B. Soros, and K.~O. Stanley, ``Quality diversity: A new frontier
  for evolutionary computation,'' {\em Frontiers in Robotics and AI}, vol.~3,
  p.~40, 2016.

\bibitem{conti2018improving}
E.~Conti, V.~Madhavan, F.~P. Such, J.~Lehman, K.~Stanley, and J.~Clune,
  ``Improving exploration in evolution strategies for deep reinforcement
  learning via a population of novelty-seeking agents,'' in {\em Advances in
  Neural Information Processing Systems}, pp.~5027--5038, 2018.

\bibitem{altman1999constrained}
E.~Altman, {\em Constrained Markov decision processes}, vol.~7.
\newblock CRC Press, 1999.

\bibitem{achiam2017constrained}
J.~Achiam, D.~Held, A.~Tamar, and P.~Abbeel, ``Constrained policy
  optimization,'' in {\em Proceedings of the 34th International Conference on
  Machine Learning-Volume 70}, pp.~22--31, JMLR. org, 2017.

\bibitem{chow2018lyapunov}
Y.~Chow, O.~Nachum, E.~Duenez-Guzman, and M.~Ghavamzadeh, ``A lyapunov-based
  approach to safe reinforcement learning,'' in {\em Advances in neural
  information processing systems}, pp.~8092--8101, 2018.

\bibitem{raybenchmarking}
A.~Ray, J.~Achiam, and D.~Amodei, ``Benchmarking safe exploration in deep
  reinforcement learning,'' {\em openai}, 2019.

\bibitem{sun2021safe}
H.~Sun, Z.~Xu, M.~Fang, Z.~Peng, J.~Guo, B.~Dai, and B.~Zhou, ``Safe
  exploration by solving early terminated mdp,'' {\em arXiv preprint
  arXiv:2107.04200}, 2021.

\bibitem{schulman2017proximal}
J.~Schulman, F.~Wolski, P.~Dhariwal, A.~Radford, and O.~Klimov, ``Proximal
  policy optimization algorithms,'' {\em arXiv preprint arXiv:1707.06347},
  2017.

\bibitem{schulman2015trust}
J.~Schulman, S.~Levine, P.~Abbeel, M.~Jordan, and P.~Moritz, ``Trust region
  policy optimization,'' in {\em International conference on machine learning},
  pp.~1889--1897, 2015.

\bibitem{lillicrap2015continuous}
T.~P. Lillicrap, J.~J. Hunt, A.~Pritzel, N.~Heess, T.~Erez, Y.~Tassa,
  D.~Silver, and D.~Wierstra, ``Continuous control with deep reinforcement
  learning,'' {\em arXiv preprint arXiv:1509.02971}, 2015.

\bibitem{fujimoto2018addressing}
S.~Fujimoto, H.~Van~Hoof, and D.~Meger, ``Addressing function approximation
  error in actor-critic methods,'' {\em arXiv preprint arXiv:1802.09477}, 2018.

\bibitem{haarnoja2018soft}
T.~Haarnoja, A.~Zhou, P.~Abbeel, and S.~Levine, ``Soft actor-critic: Off-policy
  maximum entropy deep reinforcement learning with a stochastic actor,'' {\em
  arXiv preprint arXiv:1801.01290}, 2018.

\bibitem{ruschendorf1985wasserstein}
L.~R{\"u}schendorf, ``The wasserstein distance and approximation theorems,''
  {\em Probability Theory and Related Fields}, vol.~70, no.~1, pp.~117--129,
  1985.

\bibitem{villani2008optimal}
C.~Villani, {\em Optimal transport: old and new}, vol.~338.
\newblock Springer Science \& Business Media, 2008.

\bibitem{pmlr-v70-arjovsky17a}
M.~Arjovsky, S.~Chintala, and L.~Bottou, ``{W}asserstein generative adversarial
  networks,'' in {\em Proceedings of the 34th International Conference on
  Machine Learning} (D.~Precup and Y.~W. Teh, eds.), vol.~70 of {\em
  Proceedings of Machine Learning Research}, (International Convention Centre,
  Sydney, Australia), pp.~214--223, PMLR, 06--11 Aug 2017.

\bibitem{endres2003new}
D.~M. Endres and J.~E. Schindelin, ``A new metric for probability
  distributions,'' {\em IEEE Transactions on Information theory}, 2003.

\bibitem{fuglede2004jensen}
B.~Fuglede and F.~Topsoe, ``Jensen-shannon divergence and hilbert space
  embedding,'' in {\em International Symposium onInformation Theory, 2004. ISIT
  2004. Proceedings.}, p.~31, IEEE, 2004.

\bibitem{henderson2018deep}
P.~Henderson, R.~Islam, P.~Bachman, J.~Pineau, D.~Precup, and D.~Meger, ``Deep
  reinforcement learning that matters,'' in {\em Thirty-Second AAAI Conference
  on Artificial Intelligence}, 2018.

\bibitem{ruszczynski1980feasible}
A.~Ruszczy{\'n}ski, ``Feasible direction methods for stochastic programming
  problems,'' {\em Mathematical Programming}, vol.~19, no.~1, pp.~220--229,
  1980.

\bibitem{herskovits1998feasible}
J.~Herskovits, ``Feasible direction interior-point technique for nonlinear
  optimization,'' {\em Journal of optimization theory and applications},
  vol.~99, no.~1, pp.~121--146, 1998.

\bibitem{potra2000interior}
F.~A. Potra and S.~J. Wright, ``Interior-point methods,'' {\em Journal of
  Computational and Applied Mathematics}, vol.~124, no.~1-2, pp.~281--302,
  2000.

\bibitem{dantzig2006linear}
G.~B. Dantzig and M.~N. Thapa, {\em Linear programming 2: theory and
  extensions}.
\newblock Springer Science \& Business Media, 2006.

\bibitem{conn1997globally}
A.~Conn, N.~Gould, and P.~Toint, ``A globally convergent lagrangian barrier
  algorithm for optimization with general inequality constraints and simple
  bounds,'' {\em Mathematics of Computation of the American Mathematical
  Society}, vol.~66, no.~217, pp.~261--288, 1997.

\bibitem{wright2001convergence}
S.~J. Wright, ``On the convergence of the newton/log-barrier method,'' {\em
  Mathematical Programming}, vol.~90, no.~1, pp.~71--100, 2001.

\bibitem{randlov1998learning}
J.~Randl{\o}v and P.~Alstr{\o}m, ``Learning to drive a bicycle using
  reinforcement learning and shaping.,'' in {\em ICML}, vol.~98, pp.~463--471,
  Citeseer, 1998.

\bibitem{laud2004theory}
A.~D. Laud, {\em Theory and application of reward shaping in reinforcement
  learning}.
\newblock University of Illinois at Urbana-Champaign, 2004.

\bibitem{1606.01540}
G.~Brockman, V.~Cheung, L.~Pettersson, J.~Schneider, J.~Schulman, J.~Tang, and
  W.~Zaremba, ``Openai gym,'' 2016.

\bibitem{TodorovET12}
E.~Todorov, T.~Erez, and Y.~Tassa, ``Mujoco: A physics engine for model-based
  control.,'' in {\em IROS}, pp.~5026--5033, IEEE, 2012.

\bibitem{tessler2019distributional}
C.~Tessler, G.~Tennenholtz, and S.~Mannor, ``Distributional policy
  optimization: An alternative approach for continuous control,'' {\em arXiv
  preprint arXiv:1905.09855}, 2019.

\bibitem{ciosek2019better}
K.~Ciosek, Q.~Vuong, R.~Loftin, and K.~Hofmann, ``Better exploration with
  optimistic actor critic,'' in {\em Advances in Neural Information Processing
  Systems}, pp.~1785--1796, 2019.

\end{thebibliography}
\bibliographystyle{ieeetr}

\onecolumn
\newpage
\appendix
\section{Missing Proofs}
\label{appdx:proof}
\subsection{Proof of Proposition 1}
\begin{definition}
\label{def1}
A metric space is an ordered pair $(M,d)$ where $M$ is a set and $d$ is a metric on $M$, i.e., a function
$d \colon M \times M \to \mathbb{R}$
such that for any $x,y,z\in M$, the following holds:\\
1. \quad $d(x,y)\geq 0, d(x,y)=0\Leftrightarrow x=y$, \\
2. \quad $d(x,y)=d(y,x)$, \\
3. \quad $d(x,z)\leq d(x,y)+d(y,z)$.
\end{definition}
\label{proof1}
The first two properties are obviously guaranteed by $\overline{D}_{W}^{\rho}$. As for the triangle inequality,
\begin{align*}
     &{\mathbb{E}}_{s\sim\rho(s)}[D_{W}(\theta_i(s),\theta_k(s)]\\
     &=  {\mathbb{E}}_{s\sim\rho(s)}[\sum_{l=1}^{|\mathcal{A}|}|\theta_i(s)-\theta_k(s)|] \\
    &=  {\mathbb{E}}_{s\sim\rho(s)}[\sum_{l=1}^{|\mathcal{A}|}|\theta_i(s)-\theta_j(s)+\theta_j(s)-\theta_k(s)|] \\
    &\leq  {\mathbb{E}}_{s\sim\rho(s)}[\sum_{l=1}^{(|\mathcal{A}|}|\theta_i(s)-\theta_j(s)|+|\theta_j(s)-\theta_k(s)|)]\\
    &=  {\mathbb{E}}_{s\sim\rho(s)}[\sum_{l=1}^{|\mathcal{A}|}|\theta_i(s)-\theta_j(s)|]+ {\mathbb{E}}_{s\sim\rho(s)}[\sum_{l=1}^{|\mathcal{A}|}|\theta_j(s)-\theta_k(s)|]  \\
    &= {\mathbb{E}}_{s\sim\rho(s)}[D_{W}(\theta_i(s),\theta_j(s)] +  {\mathbb{E}}_{s\sim\rho(s)}[D_{W}(\theta_j(s),\theta_k(s)] \\
\end{align*}
\section{Proof of Proposition 2}
\label{proof2}
\begin{align*}
    &\rho_{\theta}(s) = P(s_0 = s|\theta) + P(s_1 = s|\theta) + ... + P(s_T = s|\theta) \\ 
    &\quad~~\overset{L.L.N.}{=}\lim_{N\rightarrow \infty} \frac{\sum_{i=1}^N I(s_0=s|\tau_i)}{N} + \frac{\sum_{i=1}^N I(s_1=s|\tau_i)}{N} + ... + \frac{\sum_{i=1}^N I(s_T=s|\tau_i)}{N} \quad \\ 
    &\qquad~= \lim_{N\rightarrow \infty} \frac{\sum_{j=0}^T\sum_{i=1}^N I(s_j=s|\tau_i)}{N} \\
    &\overline{\rho}_{\theta}(s) = \sum_{i=1}^N\sum_{j=0}^{T} \frac{I(s_j = s|\tau_i)}{N} \\
 &\mathbb{E}[\overline{\rho}_{\theta}(s) - {\rho}_{\theta}(s)] = 0
\end{align*}

\section{Implementation Details}
\label{ImpleDetail}

\subsection{More Details on WSR}
\label{appdx_wsr}
\paragraph{WSR: Penalty Method}
The Penalty Method considers the constraints of Eq.(\ref{constr}) by putting constraint $g(\theta)$ into a penalty term, followed by solving the following unconstrained problem in an iterative manner,
\begin{equation}
    \max_{\theta\in \Theta} \quad f(\theta) + \frac{1-\alpha}{\alpha}\min\{g(\theta),0\},
\end{equation}
The limit of the above unconstrained problem when $\alpha \rightarrow 0$ then leads to the solution of the original constrained problem. As an approximation, WSR chooses a fixed weight $\alpha$, and uses the gradient of $\nabla_{\theta}f + \frac{1-\alpha}{\alpha}\nabla_{\theta}g$ instead of $\nabla_{\theta}f + \frac{1-\alpha}{\alpha}\nabla_{\theta}\min\{g(\theta),0\}$, thus the final solution will intensely rely on the selection of $\alpha$.

\subsection{Calculation of $D_{W}$}
 We use deterministic part of policies in the calculation of $D_{W}$, i.e., we remove the Gaussian noise on the action space in PPO and use $D_{W}(a_1,a_2) = |a_1 - a_2|$.

\subsection{Network Structure}
\label{network_structure}
We use MLP with 2 hidden layers as our actor models in PPO. The first hidden layer is fixed to have 32 units. We choose to use $10$, $64$ and $256$ hidden units for the three tasks respectively in all of the main experiments, after taking the success rate, performance and computation expense (i.e. the preference to use less unit when the other two factors are similar) into consideration.

\subsection{Training Timesteps}
We fix the training timesteps in our experiments. The timesteps are fixed to be $1$M in Hopper-v3, $1.6$M for Walker2d-v3 and $3$M for HalfCheetah-v3.

\section{Visualize Diversity}
\label{quali_results}

\subsection{Mujoco Locomotion}
In this section, we provide some qualitative results of IPD on the Mujoco locomotion tasks. In all of our experiments we use the vanilla Mujoco locomotion benchmarks, with the default settings on defining healthy states. Although otherwise the visualization of learned policies might become more diverse (e.g., a Hopper agent may learn to stand-up after falling down while another agent may learn to move forward on the ground if we set the $z$-axis healthy threshold as $0$).

\begin{figure}[h!]
\centering
\includegraphics[width=0.93\linewidth]{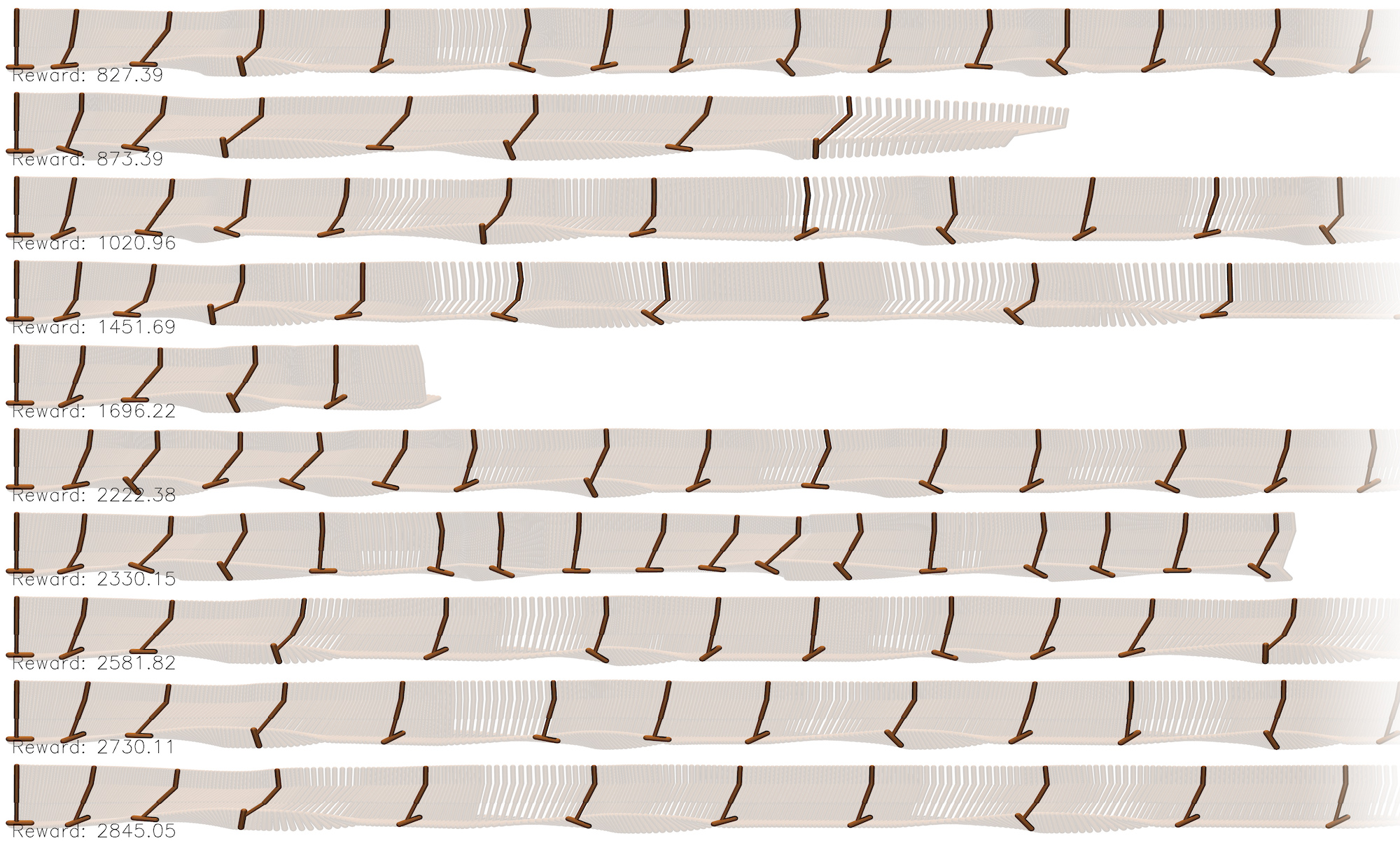}
\caption{The visualization of policy behaviors of agents trained by our method in Hopper-v3 environment. Agents learn to jump with different strides.}
\label{hopper_ours}
\end{figure}

\begin{figure}[h!]
\centering
\includegraphics[width=0.93\linewidth]{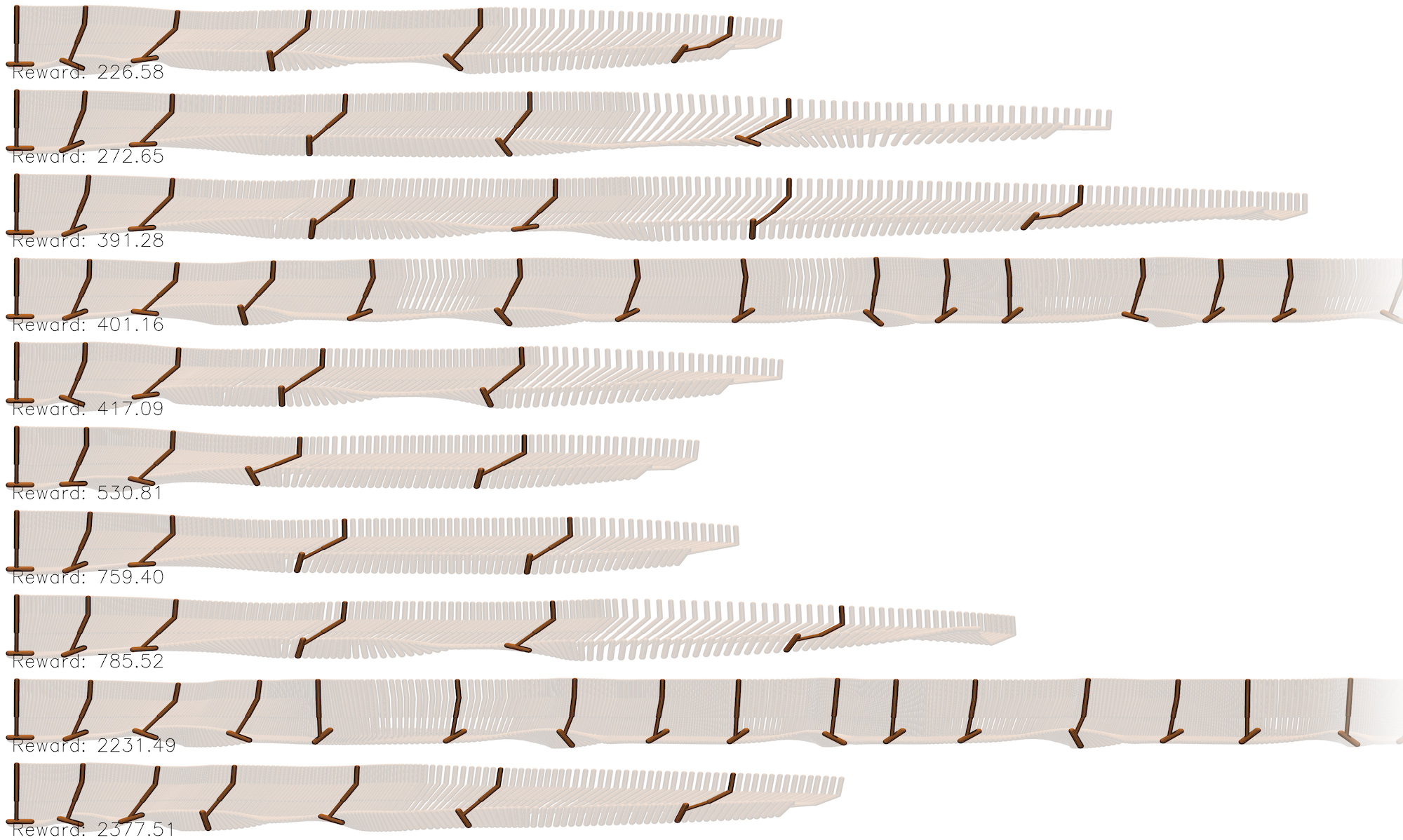}
\caption{The visualization of policy behaviors of agents trained by PPO in Hopper-v3 environment. Most agents learn a policy that can be described as \textit{Jump as far as possible and fall down}, leading to relative poor performance.}
\label{app_jump_as_far_as_possible}
\end{figure}

\begin{figure}[htb]
\centering
\includegraphics[width=\linewidth]{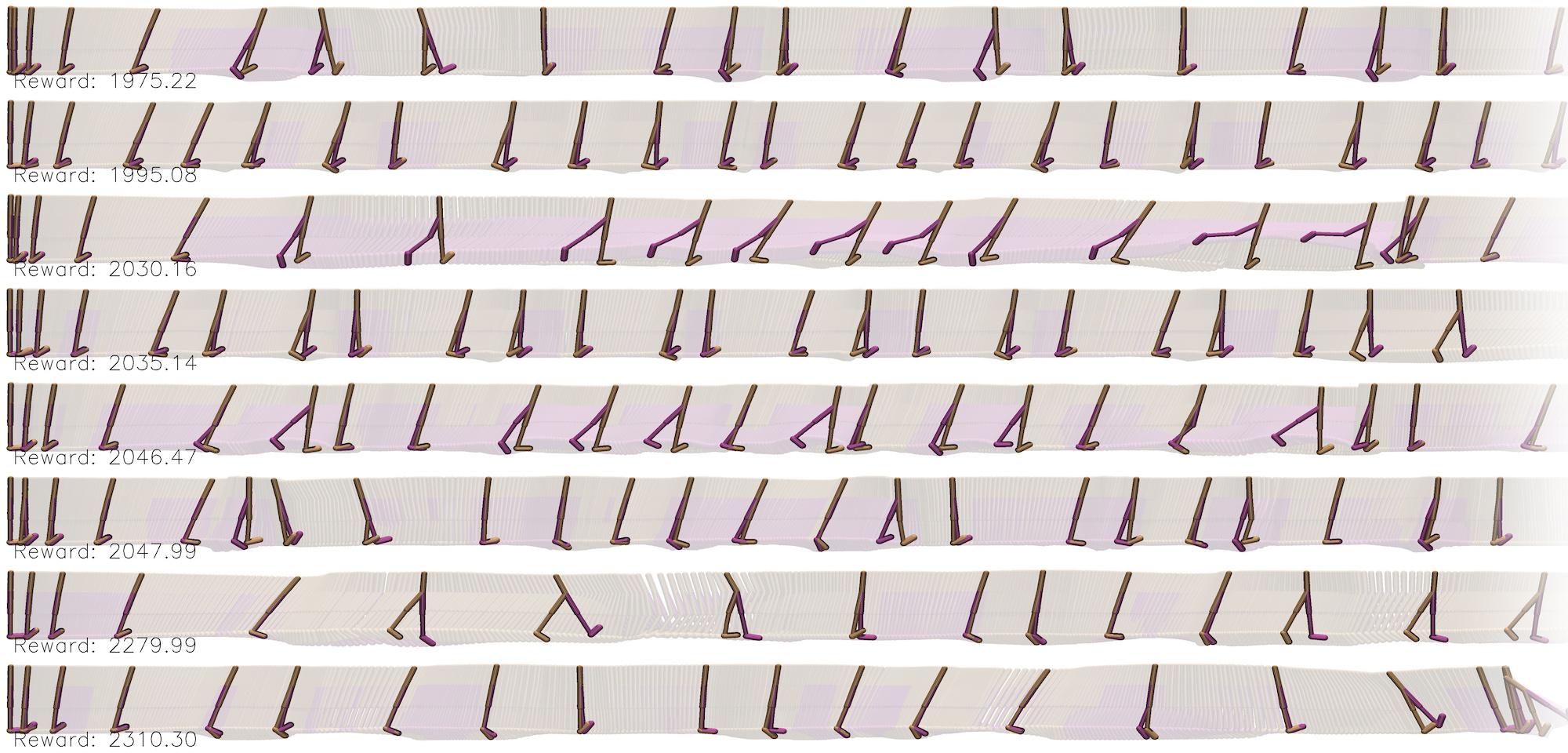}
\caption{The visualization of policy behaviors of agents trained by our method in Walker2d-v3 environment. Instead of bouncing at the ground using both legs, our agents learns to use both legs to step forward.}
\label{walker_ours}
\end{figure}

\begin{figure}[htb]
\centering
\includegraphics[width=\linewidth]{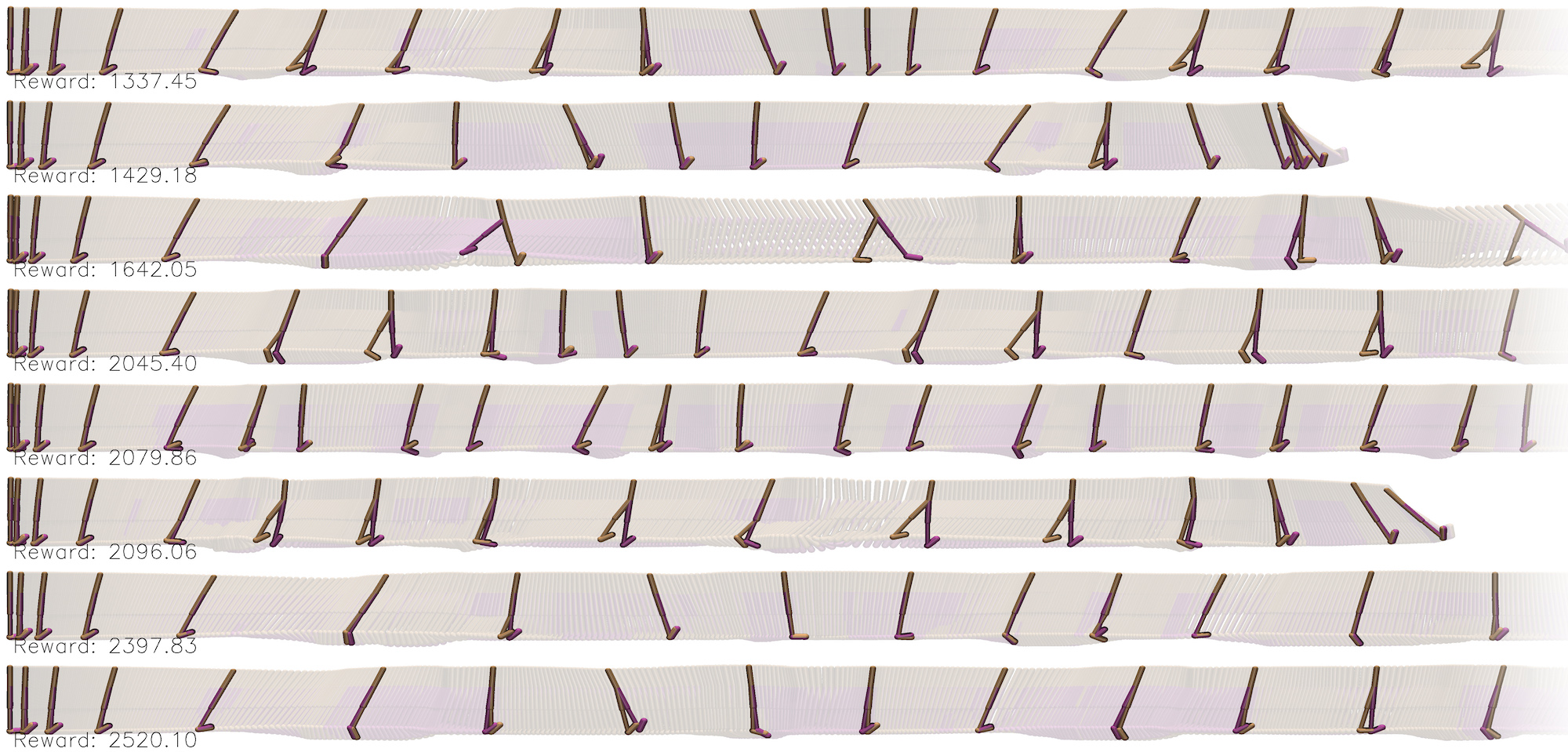}
\caption{The visualization of policy behaviors of agents trained by PPO in Walker2d-v3 environment. Most of the PPO agents only learn to use their right leg to support the body and jump forward.}
\label{walker_ppo}
\end{figure}

\begin{figure}[htb]
\centering
\includegraphics[width=0.9\linewidth]{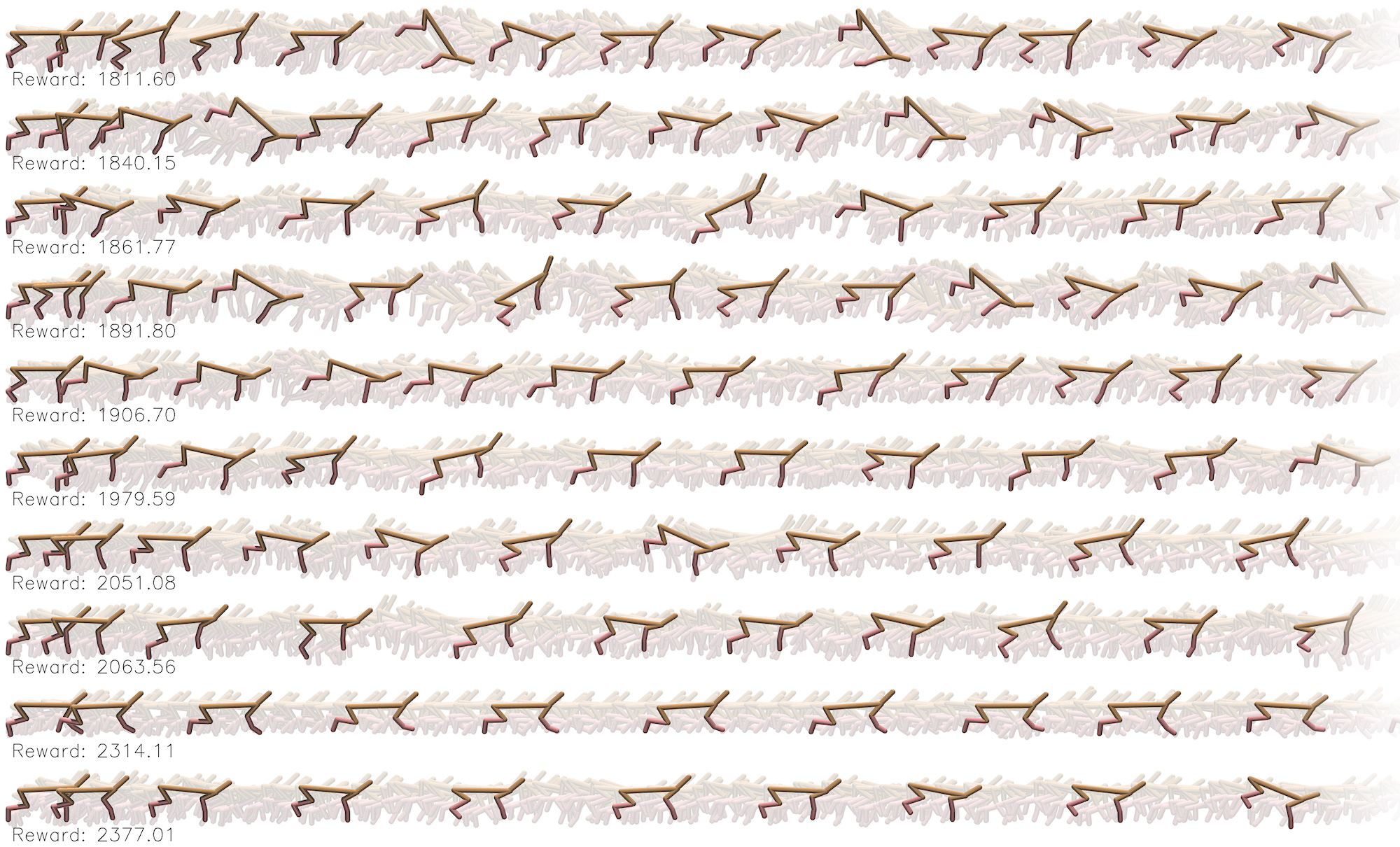}
\caption{The visualization of policy behaviors of agents trained by our method in HalfCheetah-v3 environment. Our agents run much faster compared to PPO agents and at the mean time several patterns of motion have emerged.}
\label{cheetah_ours}
\end{figure}

\begin{figure}[htb]
\centering
\includegraphics[width=0.9\linewidth]{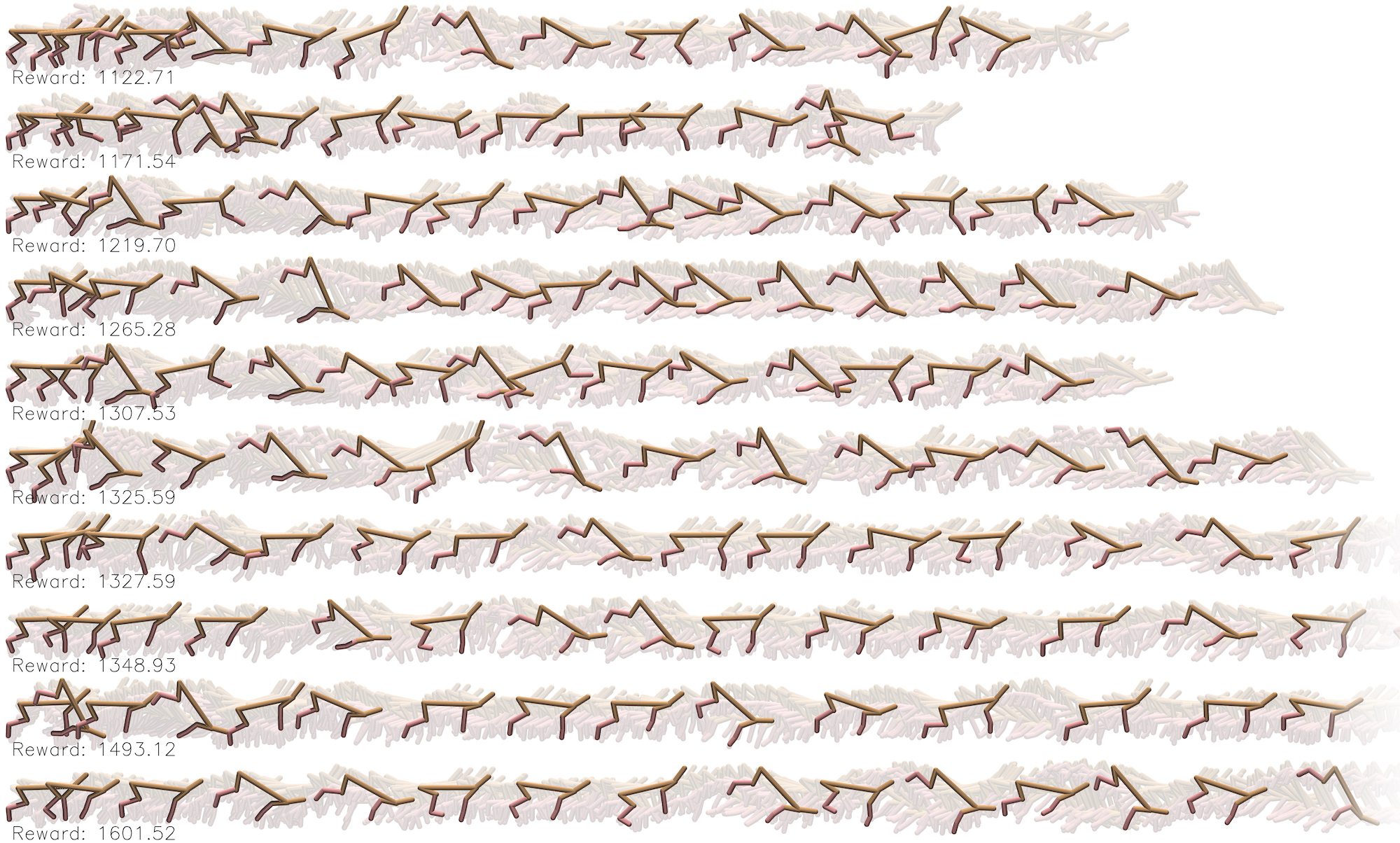}
\caption{The visualization of policy behaviors of agents trained by PPO in HalfCheetah-v3 environment. Since we only draw fixed number of frames in each line, in the limited time steps the PPO agents can not run enough distance to leave the range of our drawing, which shows that our agents run much faster.}
\label{cheetah_ppo}
\end{figure}

With the method of IPD, the Hopper policies (Figure \ref{hopper_ours}) learns to jump further and avoids falling down rather instead of just jumping and falling down (Figure \ref{app_jump_as_far_as_possible}). In the Walker2d environment, the color of purple indicates the left leg is visible. It can be seen that the IPD policies (Figure \ref{walker_ours}) learn to use both left and right legs in walking, while the PPO policies usually learn jumping. (Figure \ref{walker_ppo}). In HalfCheetah, the IPD policies (Figure \ref{cheetah_ours}) perform much better than the PPO policies (Figure \ref{cheetah_ppo}). The IPD policies leran to run with head-downward (Figure \ref{cheetah_ours} line 1), head-upward (Figure \ref{cheetah_ours} line 3), and forward (Figure \ref{cheetah_ours} line 5) while the PPO policies are always head-downward.

In Hopper and HalfCheetah, IPD is able to improve the primal task performance by avoiding always getting trapped in some certain sub-optimal behaviors.

\newpage


\end{document}